\documentclass{article}

\usepackage{microtype}
\usepackage{graphicx}
\usepackage{subcaption}
\usepackage{booktabs}
\usepackage{makecell}
\usepackage{multirow}
\usepackage{array}
\usepackage{tabularx}
\newcolumntype{Y}{>{\raggedright\arraybackslash}X}

\usepackage{hyperref}

\usepackage[accepted]{icml2026}

\usepackage{amsmath}
\usepackage{amssymb}
\usepackage{mathtools}
\usepackage{amsthm}

\usepackage[capitalize,noabbrev]{cleveref}

\theoremstyle{plain}

\theoremstyle{definition}

\theoremstyle{remark}

\usepackage[disable,textsize=tiny]{todonotes}

\icmltitlerunning{Correcting Mean Bias in Text Embeddings}

\begin{document}

\twocolumn[
\icmltitle{Correcting Mean Bias in Text Embeddings:\\A Refined Renormalization with Training-Free Improvements on MMTEB}

\icmlsetsymbol{equal}{*}

\begin{icmlauthorlist}
\icmlauthor{Xingyu Ren}{equal,cuhk}
\icmlauthor{Youran Sun}{equal,thumath}
\icmlauthor{Haoyu Liang}{tsail}
\end{icmlauthorlist}

\icmlaffiliation{cuhk}{Dept. of Physics, The Chinese University of Hong Kong, Hong Kong, China}
\icmlaffiliation{tsail}{Dept. of Computer Science and Technology, Tsinghua University, Beijing, China}
\icmlaffiliation{thumath}{Dept. of Mathematical Sciences, Tsinghua University, Beijing, China}

\icmlcorrespondingauthor{Haoyu Liang}{lianghy18@tsinghua.org.cn}

\icmlkeywords{sentence embeddings, anisotropy, mean bias, post-processing, MMTEB, mechanistic interpretability}
\vskip 0.3in
]

\printAffiliationsAndNotice{\icmlEqualContribution} 

\begin{abstract}
We find that current sentence-embedding models produce outputs with a consistent bias: every embedding $e$ decomposes as $\tilde e + \mu$, where the mean $\mu$ is near-identical across all sentences. We study two training-free corrections---subtracting $\mu$ directly (R1), or projecting each embedding off the mean direction (R2)---and show, via a first-order error-propagation argument, that R2 cancels the parallel component of mean-estimation error that R1 retains. Across 38 models on the Massive Multilingual Text Embedding Benchmark (MMTEB)~\citep{MMTEB}, R2 yields consistent classification gains (paired $\bar t = 3.31$, 29 of 38 models with $t>2$, zero losses), and the per-model mean norm $\Vert\mu\Vert$ correlates with which models benefit most. A nine-method dose-response ablation on five models further reveals that mild single-direction removal helps, but full principal component analysis (PCA) whitening hurts every model we test, and that R2 and All-but-the-Top with depth one agree within $0.18$ pp downstream despite weak geometric alignment between $\hat\mu$ and the centered top principal component.
\end{abstract}

\section{Introduction}\label{sec:intro}

Text representation collapse (embedding collapse) has recently received increasing attention. 
The cone effect has been observed in language models~\citep{ethayarajh2019contextual,gao2019representation,cai2021isotropy,liu2026dispersion} and CLIPs~\citep{liang2022mind,yi2025decipher}, and has been shown to be inherent to the self-attention mechanism itself~\citep{anisotropyInherent2024}, where text embeddings cluster within a narrow cone in high-dimensional space. 
A similar phenomenon has also been observed in text embedding models~\citep{liang2025jailbreaking}, where the embeddings are concentrated, more precisely, near the boundary of the cone.
This phenomenon leaves most regions of the representation space underutilized, undermining the model’s expressive power and robustness~\citep{liang2025jailbreaking,li2024zeroshotadversarial}.

Apply any modern sentence-embedding model $\mathcal E$ to a large diverse corpus, average the outputs, and the result is far from zero. We find that that the mean embedding
$\mu \;=\; \mathbb E_{t\sim\mathcal D}\,[\mathcal E(t)]$
is large, model-specific, and nearly unchanged across languages and prompts, according to the experimental evidence in Appendix~\ref{app:probes}.
This suggests that the bias $\mu$ is an intrinsic property of the text embedding model.


We further study whether removing this shared component, training-free, improves downstream performance, and \emph{when} it does so. These post-processing corrections sit within a long lineage of fixes for anisotropic representations~\citep{SimpleBaseline-2017,mu2017allbutthetop,BERTFlow-2020,WhiteningSentence-2021,WhiteningBERT-2021,Ditto-2023,IsAnisotropy-2022,IsotropyClusters-2024}. Let $\hat\mu = \mu/\Vert\mu\Vert$. Two natural variants are: (\textbf{R1}) subtract $\mu$ and renormalize; 
(\textbf{R2}) remove the projection of $e$ onto $\hat\mu$ and renormalize:
\begin{equation}
\text{R1:}~~~ e' = \frac{e-\mu}{\Vert e-\mu\Vert},\qquad\text{R2:}~~~ e' = \frac{e-(e\cdot\hat\mu)\hat\mu}{\Vert e-(e\cdot\hat\mu)\hat\mu\Vert}.
\label{eq:r1r2}
\end{equation}
Geometrically, R2 removes only the component of $e$ along $\hat\mu$ before renormalizing. R2 is also closely related to the All-but-the-Top (ABTT) baseline of \citet{mu2017allbutthetop}, which removes the top principal components of the centered embedding distribution. The two coincide exactly when the dominant direction of variation in centered embeddings is $\hat\mu$ itself; in practice, as we will show, the two methods are nearly indistinguishable downstream even when they are geometrically distinct.

We test three falsifiable hypotheses on the Massive Multilingual Text Embedding Benchmark (MMTEB)~\citep{MMTEB,MTEB}:
\begin{itemize}
\item \textbf{H1 (R2 $>$ R1).} R2 outperforms R1, as predicted by the error-propagation argument.
\item \textbf{H2 ($\Vert\mu\Vert$ predicts benefit).} The per-model classification benefit of R2 grows with $\Vert\mu\Vert$.
\item \textbf{H3 (Goldilocks gradient).} More aggressive anisotropy removal is \emph{not} monotonically better; full principal component analysis (PCA) whitening will not dominate single-direction removal.
\end{itemize}
We treat $\Vert\mu\Vert$ as a \emph{post-hoc} diagnostic throughout---the panel was expanded toward high-$\Vert\mu\Vert$ models after early observations, not prospectively calibrated.

Our contributions are:
\begin{enumerate}
\item A first-order error-propagation argument (Appendix~\ref{app:proof}) predicting R2 $>$ R1, confirmed across 38 MMTEB models: R2 wins retrieval in $35/38$ models, and classification in $29/38$ with zero losses.
\item $\Vert\mu\Vert$ as a cheap, post-hoc diagnostic: per-model classification $t$ correlates with $\Vert\mu\Vert$ at Pearson $r=0.72$ (full panel, $95\%$ CI $[+0.47,+0.86]$).
\item A nine-method dose-response ablation on five models ($\Vert\mu\Vert\in[0.19,0.85]$): single-direction removal helps, but full PCA whitening hurts every model we test ($\Delta\in[-5.18,-0.64]$ pp). R2 and ABTT-$1$ agree within $0.18$ pp downstream despite $\cos(\hat\mu,\text{PC1}_{\text{centered}})\in[0.03,0.51]$.
\end{enumerate}

\paragraph{Relevance to mechanistic interpretability.} We characterize $\mu$ as an output-space feature---we do not localize it to internal circuits. However, two findings are relevant to future mechanistic accounts: (i)~$\hat\mu$ is largely invariant across languages and prompts (Appendix~\ref{app:probes}), so it is a coherent encoder feature rather than a sampling artifact; (ii)~the non-monotone dose-response (Section~\ref{sec:nine}) is consistent with prior work~\citep{IsotropyClusters-2024} arguing that some anisotropy preserves useful structure: if all variance were harmful, whitening would dominate single-direction removal. Section~\ref{sec:limit} discusses limitations.

\section{Method}\label{sec:method}

\paragraph{Notation.} Let $\mathcal E$ be a unit-norm text-embedding model with output $e \in \mathbb R^d$, $\Vert e\Vert=1$ for almost all inputs (we verify this for $28/29$ models we audit; see Appendix~\ref{app:norm}). For a corpus $\{t_i\}_{i=1}^N$ we estimate the mean $\mu = N^{-1}\sum_i \mathcal E(t_i)$ and write $\hat\mu = \mu/\Vert\mu\Vert$. Throughout this paper we estimate $\hat\mu$ from $N=10^5$ English Wikipedia sentences (snapshot \texttt{20220301.en}) with character lengths in $[64, 512]$; the same corpus is used to fit centered PCA components for the ABTT and whitening baselines, with no task labels.

\paragraph{R2 dominates R1 under estimation noise (sketch).}
We always estimate $\mu$ from a finite sample, so the estimator carries an error $\epsilon$ that decomposes orthogonally with respect to the true mean direction:
$\epsilon = \epsilon_\parallel + \epsilon_\perp$, with $\epsilon_\parallel = (\epsilon\cdot\hat\mu)\hat\mu$. Under the high-dimensional near-orthogonality assumption $\tilde e\cdot\mu\approx 0$ and for $\Vert\epsilon\Vert\ll\Vert\mu\Vert$, a first-order expansion (Appendix~\ref{app:proof}) yields
\begin{equation}
\tilde e_1 \;=\; \tilde e - \epsilon_\parallel - \epsilon_\perp,\qquad
\tilde e_2 \;\approx\; \tilde e - \epsilon_\perp.
\label{eq:r2-error}
\end{equation}
Up to leading order, R2 cancels $\epsilon_\parallel$ while R1 retains both components. The argument predicts lower first-order sensitivity of R2 to parallel mean-estimation error; the size and source of downstream gains remain empirical.

\paragraph{Effect-size statistic and aggregation.}
For a (model, family) cell with $n$ tasks we report the conventional paired $t$-statistic over per-task deltas as our primary metric. We use $|t|>2$ as the win/tie/loss (W/T/L) threshold in tables; this is a per-cell effect-size window, not a Bonferroni-controlled significance level. Family-level aggregates take the unweighted mean over models with at least five tasks in the family. The $38$-model audit and the $9$-method ablation use different MMTEB snapshots and disjoint task-tracking, so per-row deltas in Table~\ref{tab:9method} should be compared only within row, not numerically pooled with Table~\ref{tab:38model}. (Appendix~\ref{app:sigma_cell} discusses an amplified secondary metric $\sigma_{\text{cell}}$ that we report alongside the conventional $t$ in supplementary tables.)

\begin{table*}[t]
\centering
\caption{R2 effect across $38$ MMTEB models, by task family. Paired $t$ is the average of per-(model, family) cell $t$-statistics; W/T/L counts how many models pass $t>2$, sit within $\pm 2$, or fall below $-2$. \textbf{Classification (bold row) is the cleanest signal and survives removing the Snowflake family.} Retrieval is positive but Snowflake-dominated.}
\label{tab:38model}
\begin{footnotesize}
\begin{tabular}{lrcccc}
\toprule
       &        & \multicolumn{2}{c}{Full panel ($38$)} & \multicolumn{2}{c}{Excluding Snowflake ($33$)} \\
\cmidrule(lr){3-4}\cmidrule(lr){5-6}
Family & \#tasks & $\bar t$ & W/T/L & $\bar t$ & W/T/L \\
\midrule
Retrieval         & 182  & $+2.99$ & 22/12/4 & $+1.99$ & 17/12/4 \\
\textbf{Classification} & \textbf{323} & $\boldsymbol{+3.31}$ & \textbf{29/9/0} & $\boldsymbol{+3.25}$ & \textbf{24/9/0} \\
Other             & 119  & $+1.55$ & 12/26/0 & $+0.85$ & \phantom{0}7/26/0 \\
\bottomrule
\end{tabular}
\end{footnotesize}
\end{table*}

\section{Main Results: 38-Model Audit}\label{sec:main}

We evaluate R1 and R2 on $38$ embedding models from $15$ organizations on MMTEB \citep{MMTEB, MTEB}, an earlier snapshot, with $609$ to $624$ tasks per model after filtering for modality mismatch / failures / timeouts (full task list in Appendix~\ref{app:filtered}). We do not run ABTT-$1$ on the full $38$-model panel; the headline classification numbers and the $\Vert\mu\Vert$-correlation below should therefore be read as evidence for \emph{single-direction correction} (whether implemented as R2 or as ABTT-$1$) relative to the unprocessed control and to R1, not as a claim that R2 outperforms a well-implemented baseline at scale. The head-to-head R2 vs ABTT-$1$ comparison within a $5$-model intersection is reported in Table~\ref{tab:9method}. Per-task scores are paired before and after applying the post-processing. Because the panel was preferentially expanded toward $\Vert\mu\Vert\geq 0.5$ after early observations, the win counts below reflect performance in the intended high-$\Vert\mu\Vert$ regime, not an unbiased estimate over the population of MMTEB-leaderboard embedding models.

\paragraph{Aggregate effects (Table~\ref{tab:38model}).} R2 produces consistent gains in classification: paired $\bar t = 3.31$ on the full $38$-model panel, with $29$ models passing $t>2$ and zero losses. The result is robust to family removal: dropping the five Snowflake-arctic-embed models leaves $\bar t = 3.25$ over the remaining $33$ models with $24$ wins and zero losses. Retrieval is more positive on the full panel ($\bar t = 2.99$) but is family-sensitive, dropping to $\bar t=1.99$ when Snowflake is removed; we therefore report retrieval as a secondary, family-sensitive finding rather than a headline. R2 outperforms R1 on retrieval in $35$ of $38$ models and on classification in $23$ of $38$ models, consistent with the outcome-level prediction of H1. Within the $5$-model intersection of Table~\ref{tab:9method}, R2 and ABTT-$1$ agree within $0.18$ pp on every model, suggesting a similar behavioral footprint rather than a quantitative dominance claim; we position R2 as a transparent, single-vector, PCA-free implementation of single-direction correction with an explicit error-propagation handle (Eq.~\ref{eq:r2-error}) and a one-scalar $\Vert\mu\Vert$ diagnostic.

\begin{figure}[t]
\centering
\includegraphics[width=\columnwidth]{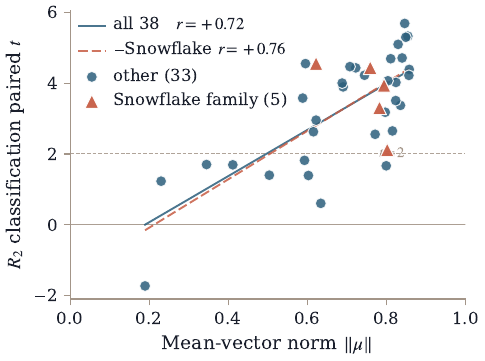}
\caption{Mean-vector norm $\Vert\mu\Vert$ vs.\ R2 classification paired $t$, $38$ MMTEB models. Pearson $r=+0.72$ on the full panel ($95\%$ bootstrap CI $[+0.47,+0.86]$) and $r=+0.76$ excluding Snowflake (CI $[+0.53,+0.88]$); Spearman $\rho=+0.61/+0.69$. Snowflake-arctic-embed models are marked separately (red triangles). The dashed line marks the $t=2$ threshold; OLS lines are descriptive. Models above the dashed line are classification ``wins'' under R2.}
\label{fig:norm_vs_t}
\vspace{-1em}
\end{figure}

\paragraph{$\Vert\mu\Vert$ correlates with classification benefit (Figure~\ref{fig:norm_vs_t}).}
Per-model classification $t$ shows a strong positive relationship with $\Vert\mu\Vert$: Pearson $r=+0.72$ on the full panel ($95\%$ bootstrap CI $[+0.47,+0.86]$) and $r=+0.76$ excluding the Snowflake family ($[+0.53,+0.88]$); Spearman $\rho = +0.61$ and $+0.69$ respectively. Across $\Vert\mu\Vert$ quartiles $[0,0.3)/[0.3,0.6)/[0.6,0.75)/[0.75,1.0)$ the mean classification $t$ grows monotonically. Computing $\Vert\mu\Vert$ from $10^5$ sentences is cheap relative to MMTEB evaluation (one forward pass over a fixed corpus, parallelizable), so the quantity is a candidate pre-deployment diagnostic; the high-$\Vert\mu\Vert$-biased panel cannot estimate the false-positive or false-negative rate of any threshold rule. We emphasize that the $\Vert\mu\Vert\geq 0.5$ regime is identified \emph{post hoc}; we have not held out a separate set of models on which to validate a prospective threshold (H2).

\paragraph{Panel-selection caveat.}
The $38$-model panel is not a uniform sample. After early observations suggested that classification benefit correlates with $\Vert\mu\Vert$, we preferentially added MMTEB-leaderboard models with $\Vert\mu\Vert\geq 0.5$, which biases the panel toward the helpful regime. The single deliberately-included low-$\Vert\mu\Vert$ model is \texttt{all-MiniLM-L6-v2} ($\Vert\mu\Vert=0.19$). We disclose this upfront and report Snowflake-free numbers throughout. The inflation also bears on the retrieval headline: roughly four-fifths of the cumulative retrieval effect on the panel is contributed by the five Snowflake models (Appendix~\ref{app:snow_share}).

\section{Goldilocks Ablation: Nine-Method Comparison}\label{sec:nine}

\begin{table*}[t]
\centering
\caption{Nine-method dose-response ladder: $\Delta$ in percentage points (mean score on the per-model intersection of tasks where all methods completed, vs.\ control). Methods: R1 (subtract $\mu$, renorm.); R2 (project out $\hat\mu$, renorm.); ABTT-$D$ (center, remove top-$D$ centered PCs, renorm.) \citep{mu2017allbutthetop}; mc (mean-centering only, no renorm.); rand (project out a single random direction, seed 42); whi (full-rank PCA whitening: center, scale each PC by $1/\sqrt{\lambda_i+\epsilon}$ with $\epsilon=10^{-6}$, renorm.). All methods estimate the necessary corpus-level statistics from the same $10^5$ Wikipedia sentences with no task labels. Per-row task counts are not numerically pooled with Table~\ref{tab:38model}.}
\label{tab:9method}
\begin{footnotesize}
\setlength{\tabcolsep}{4pt}
\begin{tabular}{lccrrrrrrrr}
\toprule
Model & $\Vert\mu\Vert$ & \#Tasks & R1 & R2 & ABTT-1 & ABTT-2 & ABTT-3 & mc & rand & whi \\
\midrule
e5-large-instr. & 0.85 & 462 & $+0.59$ & $+0.62$ & $+0.56$ & $\mathbf{+0.71}$ & $+0.58$ & $+0.07$ & $+0.00$ & $\underline{-5.18}$ \\
Snowflake-m & 0.76 & 333 & $+0.40$ & $+0.69$ & $+0.87$ & $\mathbf{+0.92}$ & $+0.78$ & $-0.14$ & $-0.02$ & $\underline{-0.64}$ \\
bge-base-v1.5 & 0.59 & 446 & $+0.22$ & $+0.23$ & $\mathbf{+0.26}$ & $+0.10$ & $+0.14$ & $+0.04$ & $-0.01$ & $\underline{-1.79}$ \\
nomic-v1 & 0.59 & 402 & $\mathbf{+0.08}$ & $+0.08$ & $+0.07$ & $+0.06$ & $-0.04$ & $-0.01$ & $-0.03$ & $\underline{-0.70}$ \\
MiniLM-L6 & 0.19 & 443 & $\mathbf{+0.03}$ & $-0.05$ & $-0.02$ & $-0.03$ & $-0.12$ & $+0.00$ & $-0.01$ & $\underline{-0.64}$ \\
\bottomrule
\end{tabular}
\end{footnotesize}
\end{table*}

To test whether more aggressive anisotropy removal is monotonically better, we run a $9$-method dose-response ladder on $5$ representative models spanning $\Vert\mu\Vert\in[0.19,0.85]$: control, R1, R2, ABTT-$\{1,2,3\}$, mean-centering only (\texttt{mc}), random-direction removal (\texttt{rand}, seed 42), and full-rank PCA whitening (\texttt{whi}). Whitening, the most aggressive entry on the ladder, centers the embeddings, scales each principal component by $1/\sqrt{\lambda_i+\epsilon}$ with $\epsilon=10^{-6}$, then $\ell_2$-renormalizes. All methods estimate the necessary corpus statistics from the same $10^5$ Wikipedia sentences with no task labels; per-row task counts in Table~\ref{tab:9method} are the per-model intersection of tasks where all $9$ methods completed (a different MMTEB snapshot from Section~\ref{sec:main}, so the rows are not numerically pooled with Table~\ref{tab:38model}).

\paragraph{Goldilocks dose-response (H3).}
Whitening, the most aggressive correction, degrades every model in the ladder, with the largest drop on the highest-$\Vert\mu\Vert$ model (\texttt{e5-large-instruct}: $-5.18$ pp). Mild single-direction removal (R2 or ABTT-$1$) instead gives positive effects on the four high-$\Vert\mu\Vert$ models ($+0.6$ to $+0.9$ pp on \texttt{Snowflake-m} and \texttt{e5-large}, $+0.2$ to $+0.3$ pp on \texttt{bge}, near-zero on \texttt{nomic}). ABTT-$2$ marginally beats R2 on the two highest-$\Vert\mu\Vert$ models, so R2 is not universally optimal, and we position it as a minimal, transparent baseline rather than a new dominant method. The dose-response is consistent with a narrow correction on the dominant bias direction, not a global isotropization.

\paragraph{R2 and ABTT-$1$: downstream-equivalent, geometrically distinct.}
R2 removes $\hat\mu$ directly; ABTT-$1$ removes the top centered PC. The cosines are surprisingly small: $0.07$ (\texttt{e5}), $0.51$ (\texttt{Snowflake}), $0.05$ (\texttt{bge}), $0.03$ (\texttt{nomic}), $0.27$ (\texttt{MiniLM}); three of five models below $0.1$ (Appendix~\ref{app:pc}). Despite this weak alignment, R2 and ABTT-$1$ agree within $0.18$ pp downstream on every model in Table~\ref{tab:9method}. The improvement is not reproduced by an arbitrary direction (Table~\ref{tab:9method}, \texttt{rand} column), but the R2/ABTT-$1$ similarity leaves open whether $\hat\mu$ is uniquely causal or one useful global direction among several, a question we cannot resolve with $5$ models and a single random-direction seed for the \texttt{rand} control.

\paragraph{Negative-direction controls.}
Mean-centering only (\texttt{mc}, no renormalization) gives $\Delta\in[-0.15,+0.07]$ pp, near-zero across the panel, ruling out the alternative that any centering operation suffices. Random-direction removal (\texttt{rand}) gives $\Delta\in[-0.03,+0.00]$ pp. To stress-test the random control we additionally ran three more seeds 2, 3, 4 on \texttt{Snowflake-m}; the four-seed mean is $+0.005$ pp (std $0.005$), versus $+0.567$ pp for R2 on the same model: random projection captures roughly $1\%$ of R2's effect on this model. The multi-seed extension covers a single model; we leave multi-model multi-seed random control to future work.

\section{Discussion and Limitations}\label{sec:limit}

The evidence above identifies a coherent feature of the output space: a shared mean direction that is large, largely invariant across language and prompt path, and whose dose-controlled removal produces a small but reliable classification gain when $\Vert\mu\Vert$ is itself large. The dose-response curve is non-monotone: single-direction removal helps in the high-$\Vert\mu\Vert$ regime, but full PCA whitening hurts every model we test, consistent with prior reports that some anisotropy preserves useful cluster structure~\citep{IsotropyClusters-2024}. Whether $\mu$ corresponds to an identifiable internal mechanism or a training-time inductive bias is an open question we do not address.

\paragraph{Limitations.} \emph{(i) Panel-selection bias.} The $38$-model panel was preferentially expanded toward $\Vert\mu\Vert\geq 0.5$ after early observations and is not a uniform sample. \emph{(ii) Snowflake dominance in retrieval.} Retrieval $\bar t$ drops $2.99\to1.99$ when the Snowflake family is removed; we treat retrieval as secondary. \emph{(iii) Post-hoc diagnostic.} The $\Vert\mu\Vert\geq 0.5$ regime is identified after observing the data, not prospectively validated on held-out models. \emph{(iv) Staged experimental rollout.} The $38$-model audit and the $9$-method ablation use different MMTEB snapshots; numbers across studies are not pooled. \emph{(v) Output-space, not circuit-level.} We do not localize $\mu$ to internal circuits or training-time causes. \emph{(vi) Single-seed random control on most models.} The four-seed random extension covers only one model. \emph{(vii) One whitening variant.} Task-adaptive whitening is out of scope. \emph{(viii) ABTT-$2$ optimality at top end and ABTT-$1$ omission from the $38$-model panel.} ABTT-$2$ beats R2 by $0.09$ pp on \texttt{e5-large-instruct} and $0.23$ pp on \texttt{Snowflake-m}; R2 is a minimal baseline. The $38$-model audit predates the ABTT comparison; the $5$-model evidence (Table~\ref{tab:9method}) shows ABTT-$1$ is downstream-equivalent within $0.18$ pp.

\paragraph{Recommendation.} On the audited panel, the $\Vert\mu\Vert\geq 0.5$ regime was associated with R2 classification gains of $+0.13$ to $+0.76$ pp on the four named models in the dose-response ladder. We do not recommend untuned full-rank PCA whitening as a default.

\bibliography{example_paper}
\bibliographystyle{icml2026}

\newpage
\appendix
\onecolumn

\section{Error-Propagation Derivation}\label{app:proof}

Let $\tilde e = e - \mu$ be the true centered embedding. Decompose the mean-estimation error as $\epsilon = \epsilon_\parallel + \epsilon_\perp$ with $\epsilon_\parallel = \alpha\,\hat\mu$ where $\alpha = \epsilon\cdot\hat\mu$ is a signed scalar, and $\epsilon_\perp\cdot\mu = 0$. Assume $\tilde e\cdot\mu\approx 0$ and $\tilde e\cdot\epsilon\approx 0$ (high-dimensional near-orthogonality of signal and bias direction). Note that $\alpha$ may be of either sign; the parallel error can over- or under-estimate $\Vert\mu\Vert$.

\paragraph{R1.} Subtracting the estimated mean gives
\begin{equation}
\tilde e_1 = e - (\mu + \epsilon) = \tilde e - \alpha\,\hat\mu - \epsilon_\perp,
\end{equation}
which retains both the parallel and orthogonal components of $\epsilon$.

\paragraph{R2.} Projecting out the estimated mean direction gives
\begin{equation}
\tilde e_2 = e - \frac{e\cdot(\mu+\epsilon)}{\Vert\mu+\epsilon\Vert^2}\,(\mu+\epsilon).
\end{equation}
We expand numerator and denominator to first order in $\Vert\epsilon\Vert/\Vert\mu\Vert$. Using $\tilde e\cdot\mu\approx 0$ and $\tilde e\cdot\epsilon\approx 0$:
\begin{equation}
e\cdot(\mu+\epsilon) = (\tilde e + \mu)\cdot(\mu + \alpha\hat\mu + \epsilon_\perp) \approx \Vert\mu\Vert^2 + \alpha\Vert\mu\Vert,
\end{equation}
\begin{equation}
\Vert\mu+\epsilon\Vert^2 = (\Vert\mu\Vert + \alpha)^2 + \Vert\epsilon_\perp\Vert^2 \approx \Vert\mu\Vert^2 + 2\alpha\Vert\mu\Vert + O(\Vert\epsilon\Vert^2).
\end{equation}
The ratio simplifies to first order as
\begin{equation}
\frac{\Vert\mu\Vert^2 + \alpha\Vert\mu\Vert}{\Vert\mu\Vert^2 + 2\alpha\Vert\mu\Vert} \approx 1 - \frac{\alpha}{\Vert\mu\Vert} + O\!\left(\frac{\alpha^2}{\Vert\mu\Vert^2}\right).
\end{equation}
Substituting,
\begin{equation}
\tilde e_2 \approx \tilde e + \mu - \left(1 - \tfrac{\alpha}{\Vert\mu\Vert}\right)(\mu + \alpha\hat\mu + \epsilon_\perp).
\end{equation}
Expanding the product, the leading parallel cancellation reads
\begin{equation}
\mu - \left(1 - \tfrac{\alpha}{\Vert\mu\Vert}\right)\mu = \alpha\,\hat\mu, \qquad - \left(1 - \tfrac{\alpha}{\Vert\mu\Vert}\right)\alpha\,\hat\mu \approx -\alpha\,\hat\mu,
\end{equation}
so the two parallel contributions cancel to first order, and the orthogonal residual is
\begin{equation}
- \left(1 - \tfrac{\alpha}{\Vert\mu\Vert}\right)\epsilon_\perp \approx -\,\epsilon_\perp.
\end{equation}
Combining,
\begin{equation}
\tilde e_2 \approx \tilde e - \epsilon_\perp.
\end{equation}
R2 therefore cancels the parallel error $\alpha\,\hat\mu$ to leading order, while R1 retains it. The argument characterizes estimation-error propagation only; downstream score gains and their source remain empirical.

\paragraph{Consequence for finite-corpus estimation.} If $\hat\mu$ is estimated from $N$ Wikipedia sentences, $\alpha$ is itself a random variable with variance scaling like $1/N$. R2 makes the leading-order behaviour insensitive to this parallel component, leaving only the orthogonal residual (which has variance scaling like $(d-1)/(N\cdot\Vert\mu\Vert^2)$ in the high-dimensional regime) as the noise that survives renormalization. R1 retains $\alpha\,\hat\mu$ regardless of $N$.

\section{Effect-Size Statistic}\label{app:sigma_cell}

The body uses two metrics for a (model, family) cell with $n$ tasks. The primary metric is the conventional paired $t$-statistic $t = \bar\Delta/\mathrm{SE}(\Delta)$ over per-task deltas $\Delta_i$; this is what most reviewers expect. The secondary metric, which we call $\sigma_{\text{cell}}$, is
\begin{equation}
\sigma_{\text{cell}} \;=\; \frac{n\,\bar\Delta}{s_{\text{baseline}}}, \qquad s_{\text{baseline}} = \sqrt{\tfrac{1}{n-1}\sum_i (s_i^{\text{ctrl}}-\bar s^{\text{ctrl}})^2}.
\end{equation}
The two are related by $\sigma_{\text{cell}} = \sqrt{n}\,\bar\Delta / \sigma^{\text{SE}}$ where $\sigma^{\text{SE}}=s_{\text{baseline}}/\sqrt{n}$ is the standard error of the baseline-score mean. $\sigma_{\text{cell}}$ amplifies effects in cells with many baseline-similar tasks (large $n$, small $s_{\text{baseline}}$) and is therefore not a paired test. We report it because it is informative about per-cell coherence (it amplifies signals in families with many homogeneous tasks); the W/T/L thresholds at $|\sigma_{\text{cell}}|>2$ in our tables are an internal effect-size window, not a Bonferroni-controlled significance level. Throughout this paper, the conventional paired $t$ is the primary metric for any claim about whether R2 helps or hurts.

\section{Embedding-Norm Audit}\label{app:norm}

Each evaluated model is wrapped via the standard \texttt{SentenceTransformer} call path used in evaluation. We re-encoded $200$ sentences per model and measured $\Vert e\Vert_2$ per output. Of $29$ models that loaded successfully on our audit environment, $28$ produce unit-norm embeddings ($\bar{\Vert e\Vert} = 1.0000$, std $\le 10^{-3}$). The single non-unit model is \texttt{paraphrase-multilingual-MiniLM-L12-v2} (mean $\approx 4.3$). The $9$ audit failures stem from BFloat16-related dtype issues or custom modeling code that bypasses the audit harness; all $9$ models were nonetheless successfully evaluated end-to-end in the primary pipeline. This audit rules out a normalization confound for R1, R2, and ABTT on the remaining $28$ models. Raw per-model statistics accompany this submission.

\section{PC1 vs Mean Alignment: Centered and Uncentered}\label{app:pc}

There is a folklore observation that ``the first principal component of sentence embeddings is essentially the mean direction.'' This is true for the \emph{uncentered} principal direction (the right singular vector of the raw embedding matrix), because the large mean-norm $\Vert\mu\Vert$ makes $\mu\mu^\top$ dominate the uncentered second moment $\mathbb E[ee^\top] = \mathbb E[\tilde e\tilde e^\top] + \mu\mu^\top$.

The All-but-the-Top algorithm of \citet{mu2017allbutthetop} works on \emph{centered} embeddings instead: it subtracts $\mu$ first, then removes the top principal component(s) of the residual distribution. After centering, the $\mu$-direction variance is gone, and the centered top PC is whatever direction had the second-largest variance in the raw embeddings. There is no a priori reason for that residual direction to align with $\hat\mu$.

Both quantities are well-defined and easy to compute. We report both:
\begin{center}
\begin{tabular}{lccc}
\toprule
Model & $\Vert\mu\Vert$ & $\cos(\hat\mu,\,\mathrm{PC1}_{\text{centered}})$ & $\cos(\hat\mu,\,\mathrm{PC1}_{\text{uncentered}})$ \\
\midrule
intfloat/multilingual-e5-large-instruct & 0.85 & 0.0699 & 1.0000 \\
Snowflake/snowflake-arctic-embed-m      & 0.76 & 0.5149 & 0.9998 \\
BAAI/bge-base-en-v1.5                   & 0.59 & 0.0477 & 1.0000 \\
nomic-ai/nomic-embed-text-v1            & 0.60 & 0.0280 & 1.0000 \\
sentence-transformers/all-MiniLM-L6-v2  & 0.19 & 0.2663 & 0.9757 \\
\bottomrule
\end{tabular}
\end{center}

The right column recovers the folklore: on the four high-$\Vert\mu\Vert$ models, the uncentered top principal direction is indistinguishable from $\hat\mu$ to four decimal places. The exception is the low-$\Vert\mu\Vert$ \texttt{all-MiniLM-L6-v2}, where $\Vert\mu\Vert^2 \approx 1.5\,\lambda_1^{\text{centered}}$ is no longer dominant over the centered top eigenvalue and $\hat\mu$ ceases to be (nearly) the top uncentered direction.

Closed-form computation. The uncentered second-moment matrix is a rank-1 update $C = C_{\text{centered}} + \Vert\mu\Vert^2\,\hat\mu\hat\mu^\top$ of the centered covariance. Expanding $\hat\mu = \sum_i \alpha_i u_i$ in the centered eigenbasis, the largest eigenvalue $\lambda^*$ of $C$ is the largest root of the secular equation $1 = \Vert\mu\Vert^2 \sum_i \alpha_i^2 / (\lambda^* - \lambda_i^{\text{centered}})$, and the corresponding eigenvector is $v_1 \propto \sum_i [\alpha_i / (\lambda^* - \lambda_i^{\text{centered}})]\,u_i$. We solve this in float64 from the saved centered eigenvectors and eigenvalues; the closed-form solver and the resolved cosines accompany this submission, and no additional embedding forward passes were required.

\paragraph{Why this is interesting for ABTT-$1$ vs R2.} On the four high-$\Vert\mu\Vert$ models, the centered top PC is nearly orthogonal to $\hat\mu$ (cos $\le 0.07$ for three of them). ABTT-$1$ therefore performs two essentially independent operations: (i) subtract the mean direction, and (ii) remove a separate, almost-orthogonal residual direction. R2 only does step (i). Despite this geometric difference, R2 and ABTT-$1$ give almost identical downstream effects (Table~\ref{tab:9method}, within $0.18$\,pp on every model). The simplest reading is that step (ii) of ABTT-$1$ contributes very little downstream on these models, and the dominant effect comes from removing the $\hat\mu$ direction, which both methods do. We phrase this as a hypothesis rather than a proven mechanism. On the low-$\Vert\mu\Vert$ MiniLM the picture is different: there is non-trivial $\hat\mu$/centered-PC1 alignment ($0.27$), so step (ii) partially overlaps with step (i); both methods give near-zero downstream effect, consistent with the small $\Vert\mu\Vert$ leaving little anisotropy to remove in the first place.

\section{Multilingual and Prompt-Path Probes}\label{app:probes}

\paragraph{Multilingual $\mu$.} For \texttt{multilingual-e5-large-instruct}, we recomputed $\mu$ using $10$ Wikipedia language editions ($10\,000$ sentences each: en, de, fr, es, zh, ja, ru, ar, ko, pt). The running mean stabilizes after the first few languages:
\begin{center}
\begin{tabular}{lcc}
\toprule
After language & running $\Vert\mu\Vert$ & cumulative samples \\
\midrule
english (en, init.) & 0.8496 & 10\,000 \\
+ german (de) & 0.8507 & 20\,000 \\
+ french (fr) & 0.8508 & 30\,000 \\
+ spanish (es) & 0.8497 & 40\,000 \\
+ chinese (zh) & 0.8463 & 50\,000 \\
+ japanese (ja) & 0.8453 & 60\,000 \\
+ russian (ru) & 0.8445 & 70\,000 \\
+ arabic (ar) & 0.8427 & 80\,000 \\
+ korean (ko) & 0.8424 & 90\,000 \\
+ portuguese (pt, final) & 0.8431 & 100\,000 \\
\bottomrule
\end{tabular}
\end{center}
Final norms: $\Vert\mu_{\text{en}}\Vert=0.849$, $\Vert\mu_{\text{multi}}\Vert=0.843$; $\cos(\mu_{\text{en}},\mu_{\text{multi}})=0.973$ (angle $13.2^\circ$). The bias direction is largely model-intrinsic rather than language-specific. This does \emph{not} constitute a downstream multilingual re-evaluation, which we leave to future work.

\paragraph{Prompt-path probe.} For the same instruct model we recomputed $\mu$ on $10^5$ sentences using two representative prompts:
\begin{itemize}
\item \emph{Retrieval prompt}: ``Instruct: Given a web search query, retrieve relevant passages that answer the query\textbackslash nQuery: ''
\item \emph{Classification prompt}: ``Instruct: Classify the following text\textbackslash nQuery: ''
\end{itemize}
Norms and inter-prompt cosines:
\begin{center}
\begin{tabular}{lccc}
\toprule
& plain & retrieval prompt & classification prompt \\
\midrule
$\Vert\mu\Vert$ & 0.849 & 0.861 & 0.926 \\
\midrule
plain &  & 0.991 (7.8$^\circ$) & 0.868 (29.7$^\circ$) \\
retrieval prompt &  &  & 0.840 (32.9$^\circ$) \\
\bottomrule
\end{tabular}
\end{center}
Prompt mismatch is minor for retrieval-style prompts and moderate for classification-style prompts. We did not re-run R2 with prompted $\hat\mu$.

\section{Multi-Seed Random-Direction Control}\label{app:seeds}

On \texttt{Snowflake-arctic-embed-m} we ran three additional random-direction seeds beyond the seed-$42$ \texttt{random} column of Table~\ref{tab:9method}.
\begin{center}
\begin{tabular}{lcc}
\toprule
Seed & $\Delta$ (pp) on per-task intersection & ratio to R2 \\
\midrule
42 & $+0.0015$ & $0.3\%$ \\
2  & $+0.0009$ & $0.2\%$ \\
3  & $+0.0072$ & $1.3\%$ \\
4  & $+0.0106$ & $1.9\%$ \\
\midrule
mean & $+0.005$ (std $0.005$) & $0.9\%$ \\
\midrule
R2 (reference) & $+0.567$ &  \\
\bottomrule
\end{tabular}
\end{center}
The mean random-direction effect captures roughly $1\%$ of the R2 effect on this model. Multi-model multi-seed extension is left to future work.

\section{$\mu$-Decomposition Schematic}\label{app:figure}

\begin{figure}[h]
\centering
\includegraphics[width=0.95\textwidth]{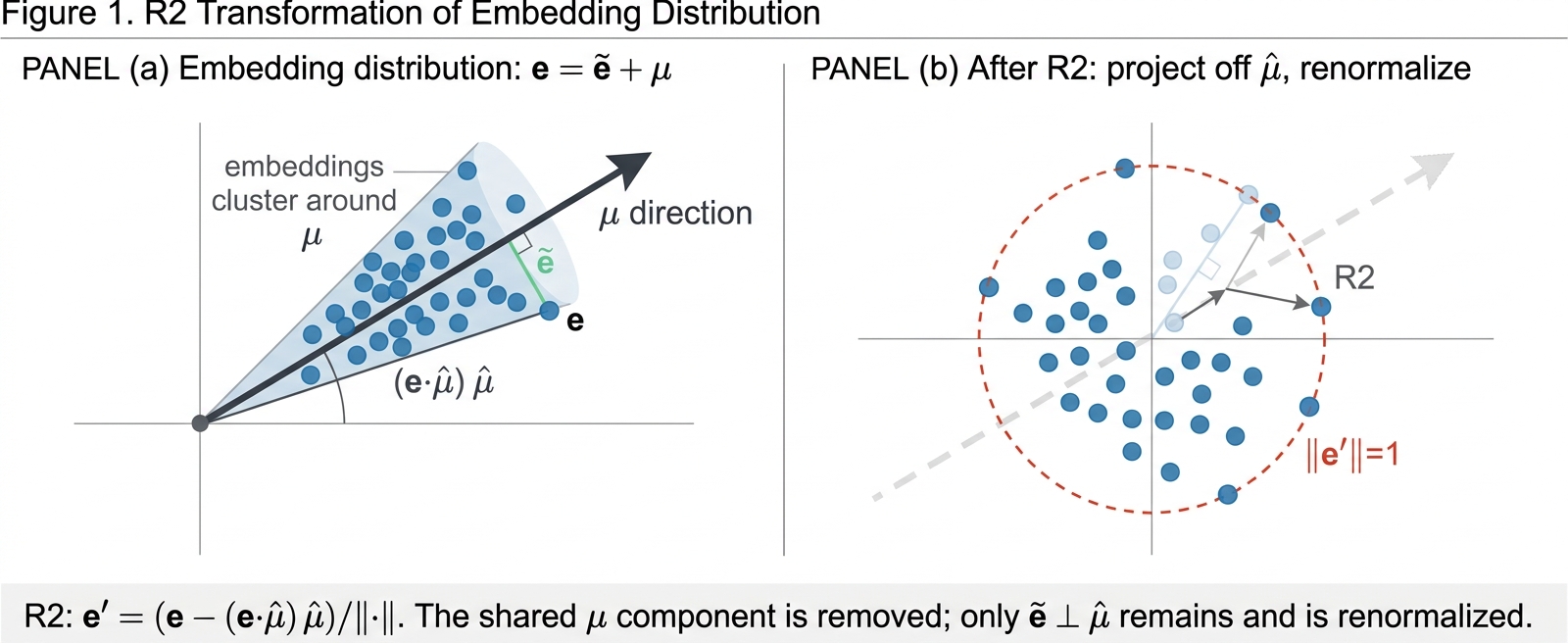}
\caption{Geometric intuition for R2. \emph{Left}: each embedding decomposes as $e=\tilde e+\mu$, and embeddings concentrate around the mean direction $\hat\mu$. \emph{Right}: R2 projects $e$ off $\hat\mu$ and renormalizes to unit length, leaving $\tilde e_2 \approx \tilde e - \epsilon_\perp$ to first order in the mean-estimation error.}
\label{fig:mu_decomp}
\end{figure}

\section{$\Vert\mu\Vert$ Quartile Stratification}\label{app:quartile}

\begin{table}[ht]
\centering
\caption{Mean R2 paired $t$-statistic per family, stratified by $\Vert\mu\Vert$ quartile.}
\label{tab:mu_quartile}
\small
\begin{tabular}{lrrrr}
\toprule
$\Vert\mu\Vert$ bin & \# & Retrieval & Classif. & Other \\
\midrule
{}[0.00,0.30) & 2 & $+2.34$ & $-0.25$ & $+0.05$ \\
{}[0.30,0.60) & 6 & $+1.36$ & $+2.46$ & $+1.23$ \\
{}[0.60,0.75) & 10 & $+2.15$ & $+3.31$ & $+1.39$ \\
{}[0.75,1.05) & 20 & $+3.95$ & $+3.91$ & $+1.88$ \\
\bottomrule
\end{tabular}
\end{table}

\noindent The classification effect grows monotonically across $\Vert\mu\Vert$ quartiles. The retrieval effect grows from quartile 1 to quartiles 3-4 but is bin-sensitive: the upper retrieval quartile is dominated by the Snowflake family (see Appendix~\ref{app:snow_share}). The Other family stays small and noisy throughout, consistent with the body's positioning of retrieval as a family-sensitive secondary finding.

\section{R2 vs R1 Retrieval Head-to-Head}\label{app:r2_v_r1}

\begin{table}[ht]
\centering
\caption{Retrieval R1-vs-R2 head-to-head on the $38$-model panel, sorted by $\Vert\mu\Vert$. R2 wins $35/38$ on paired $t$ and $36/38$ on $\sigma_{\text{cell}}$; the three exceptions are highlighted with a dagger.}
\label{tab:r2_vs_r1_retrieval}
\scriptsize
\setlength{\tabcolsep}{4pt}
\begin{tabular}{lrrrrrrcc}
\toprule
Model & $\Vert\mu\Vert$ & \#T & $t_{R_1}$ & $t_{R_2}$ & $\sigma_{R_1}$ & $\sigma_{R_2}$ & $t_{R_2}{>}t_{R_1}$ & $\sigma_{R_2}{>}\sigma_{R_1}$ \\
\midrule
\texttt{e5-small} & 0.86 & 182 & $-1.02$ & $-1.15$ & $-1.1$ & $-1.2$ & $\dagger$ & $\dagger$ \\
\texttt{rubert-tiny-turbo} & 0.86 & 179 & $+6.09$ & $+6.05$ & $+15.6$ & $+16.1$ & $\dagger$ & \checkmark \\
\texttt{multilingual-e5-small} & 0.85 & 181 & $+2.87$ & $+2.90$ & $+2.6$ & $+2.6$ & \checkmark & \checkmark \\
\texttt{multilingual-e5-large-inst..} & 0.85 & 182 & $+7.94$ & $+8.01$ & $+12.5$ & $+12.7$ & \checkmark & \checkmark \\
\texttt{e5-small-v2} & 0.85 & 182 & $+0.12$ & $+0.15$ & $+0.1$ & $+0.2$ & \checkmark & \checkmark \\
\texttt{sentence-t5-base} & 0.84 & 182 & $-4.00$ & $-3.77$ & $-2.3$ & $-2.1$ & \checkmark & \checkmark \\
\texttt{e5-base} & 0.84 & 182 & $+4.80$ & $+4.98$ & $+5.9$ & $+6.1$ & \checkmark & \checkmark \\
\texttt{e5-base-v2} & 0.83 & 174 & $+1.72$ & $+1.74$ & $+1.5$ & $+1.5$ & \checkmark & $\dagger$ \\
\texttt{granite-embedding-125m-eng..} & 0.82 & 181 & $+2.17$ & $+2.27$ & $+0.5$ & $+0.6$ & \checkmark & \checkmark \\
\texttt{bge-base-en} & 0.82 & 180 & $+9.34$ & $+9.55$ & $+8.3$ & $+8.4$ & \checkmark & \checkmark \\
\texttt{GIST-all-MiniLM-L6-v2} & 0.82 & 180 & $+6.02$ & $+6.10$ & $+4.1$ & $+4.1$ & \checkmark & \checkmark \\
\texttt{sentence-t5-large} & 0.81 & 178 & $-5.08$ & $-4.44$ & $-3.1$ & $-2.5$ & \checkmark & \checkmark \\
\texttt{LaBSE-ru-turbo} & 0.80 & 181 & $+4.79$ & $+4.86$ & $+10.6$ & $+10.8$ & \checkmark & \checkmark \\
\texttt{snowflake-arctic-embed-xs} & 0.80 & 182 & $+9.18$ & $+9.32$ & $+37.0$ & $+38.9$ & \checkmark & \checkmark \\
\texttt{rubert-tiny2} & 0.80 & 181 & $+2.46$ & $+2.87$ & $+5.2$ & $+6.1$ & \checkmark & \checkmark \\
\texttt{jina-embeddings-v2-base-en} & 0.80 & 173 & $-0.29$ & $-0.29$ & $-0.2$ & $-0.2$ & $\dagger$ & \checkmark \\
\texttt{snowflake-arctic-embed-m-l..} & 0.79 & 175 & $+8.09$ & $+8.17$ & $+43.6$ & $+46.9$ & \checkmark & \checkmark \\
\texttt{snowflake-arctic-embed-s} & 0.78 & 182 & $+10.16$ & $+10.39$ & $+47.6$ & $+51.5$ & \checkmark & \checkmark \\
\texttt{GIST-small-Embedding-v0} & 0.77 & 182 & $+1.57$ & $+1.82$ & $+0.9$ & $+1.0$ & \checkmark & \checkmark \\
\texttt{snowflake-arctic-embed-m} & 0.76 & 181 & $+9.10$ & $+9.54$ & $+74.2$ & $+90.1$ & \checkmark & \checkmark \\
\texttt{GIST-Embedding-v0} & 0.74 & 182 & $+1.52$ & $+1.63$ & $+0.6$ & $+0.7$ & \checkmark & \checkmark \\
\texttt{MedEmbed-small-v0.1} & 0.72 & 181 & $+2.59$ & $+3.24$ & $+1.4$ & $+1.7$ & \checkmark & \checkmark \\
\texttt{granite-embedding-107m-mul..} & 0.71 & 181 & $-4.26$ & $-4.11$ & $-1.6$ & $-1.5$ & \checkmark & \checkmark \\
\texttt{granite-embedding-30m-engl..} & 0.69 & 181 & $+1.63$ & $+1.65$ & $+0.7$ & $+0.8$ & \checkmark & \checkmark \\
\texttt{granite-embedding-278m-mul..} & 0.69 & 181 & $-3.21$ & $-2.35$ & $-1.1$ & $-0.9$ & \checkmark & \checkmark \\
\texttt{gte-multilingual-base} & 0.63 & 181 & $+3.25$ & $+3.59$ & $+1.7$ & $+1.9$ & \checkmark & \checkmark \\
\texttt{bge-small-en-v1.5} & 0.62 & 182 & $+5.32$ & $+5.86$ & $+2.7$ & $+2.9$ & \checkmark & \checkmark \\
\texttt{snowflake-arctic-embed-m-v..} & 0.62 & 181 & $+10.00$ & $+10.44$ & $+52.8$ & $+69.8$ & \checkmark & \checkmark \\
\texttt{Wartortle} & 0.62 & 182 & $-2.21$ & $-1.99$ & $-1.0$ & $-0.9$ & \checkmark & \checkmark \\
\texttt{paraphrase-multilingual-Mi..} & 0.60 & 181 & $+1.62$ & $+3.56$ & $+0.3$ & $+0.9$ & \checkmark & \checkmark \\
\texttt{Squirtle} & 0.60 & 182 & $-1.74$ & $-1.16$ & $-1.0$ & $-0.6$ & \checkmark & \checkmark \\
\texttt{nomic-embed-text-v1} & 0.59 & 174 & $-2.00$ & $-1.06$ & $-1.9$ & $-0.9$ & \checkmark & \checkmark \\
\texttt{bge-base-en-v1.5} & 0.59 & 181 & $+5.00$ & $+5.52$ & $+3.2$ & $+3.5$ & \checkmark & \checkmark \\
\texttt{USER-base} & 0.50 & 179 & $-1.49$ & $-0.42$ & $-1.1$ & $-0.3$ & \checkmark & \checkmark \\
\texttt{LaBSE-en-ru} & 0.41 & 182 & $-1.51$ & $+2.59$ & $-1.3$ & $+1.8$ & \checkmark & \checkmark \\
\texttt{potion-multilingual-128M} & 0.34 & 182 & $-8.90$ & $+2.70$ & $-4.8$ & $+1.2$ & \checkmark & \checkmark \\
\texttt{potion-base-8M} & 0.23 & 180 & $-3.20$ & $+1.04$ & $-1.5$ & $+0.4$ & \checkmark & \checkmark \\
\texttt{all-MiniLM-L6-v2} & 0.19 & 169 & $+3.00$ & $+3.64$ & $+0.6$ & $+0.9$ & \checkmark & \checkmark \\
\midrule
\multicolumn{7}{l}{Total: $t_{R_2}>t_{R_1}$ in 35/38 models; $\sigma_{R_2}>\sigma_{R_1}$ in 36/38 models.} \\
\bottomrule
\end{tabular}
\end{table}

\noindent Across the full $38$-model panel, R2's retrieval paired $t$-statistic exceeds R1's on $35$ of $38$ models, and R2's $\sigma_{\text{cell}}$ exceeds R1's on $36$ of $38$ models. The exceptions are flagged with $\dagger$ in the table. This pattern is consistent with the first-order error-propagation prediction (Appendix~\ref{app:proof}) but does not isolate the mechanism, since MMTEB scores cannot independently vary the parallel-vs-orthogonal split of the mean-estimation error.

\section{Snowflake Share Decomposition for Retrieval}\label{app:snow_share}

\begin{table}[ht]
\centering
\caption{Per-model R2 retrieval $\sigma_{\text{cell}}$ contribution to the panel aggregate. Total $= 372.7$; Snowflake family contributes $297.2$ (79.7\%) of the cumulative $\sigma_{\text{cell}}$. The remaining $33$ models contribute the balance. Models sorted by individual contribution.}
\label{tab:snow_share}
\scriptsize
\setlength{\tabcolsep}{6pt}
\begin{tabular}{lrrr}
\toprule
Model & $\Vert\mu\Vert$ & $\sigma_{\text{cell}}^{R_2}$ & \% of total \\
\midrule
\texttt{snowflake-arctic-embed-m} $\spadesuit$ & 0.76 & $+90.09$ & +24.2 \\
\texttt{snowflake-arctic-embed-m-v..} $\spadesuit$ & 0.62 & $+69.81$ & +18.7 \\
\texttt{snowflake-arctic-embed-s} $\spadesuit$ & 0.78 & $+51.48$ & +13.8 \\
\texttt{snowflake-arctic-embed-m-l..} $\spadesuit$ & 0.79 & $+46.88$ & +12.6 \\
\texttt{snowflake-arctic-embed-xs} $\spadesuit$ & 0.80 & $+38.94$ & +10.4 \\
\texttt{rubert-tiny-turbo}  & 0.86 & $+16.06$ & +4.3 \\
\texttt{multilingual-e5-large-inst..}  & 0.85 & $+12.65$ & +3.4 \\
\texttt{LaBSE-ru-turbo}  & 0.80 & $+10.82$ & +2.9 \\
\texttt{bge-base-en}  & 0.82 & $+8.44$ & +2.3 \\
\texttt{rubert-tiny2}  & 0.80 & $+6.11$ & +1.6 \\
\texttt{e5-base}  & 0.84 & $+6.06$ & +1.6 \\
\texttt{GIST-all-MiniLM-L6-v2}  & 0.82 & $+4.14$ & +1.1 \\
\texttt{bge-base-en-v1.5}  & 0.59 & $+3.53$ & +0.9 \\
\texttt{bge-small-en-v1.5}  & 0.62 & $+2.93$ & +0.8 \\
\texttt{multilingual-e5-small}  & 0.85 & $+2.60$ & +0.7 \\
\texttt{gte-multilingual-base}  & 0.63 & $+1.86$ & +0.5 \\
\texttt{LaBSE-en-ru}  & 0.41 & $+1.80$ & +0.5 \\
\texttt{MedEmbed-small-v0.1}  & 0.72 & $+1.69$ & +0.5 \\
\texttt{e5-base-v2}  & 0.83 & $+1.45$ & +0.4 \\
\texttt{potion-multilingual-128M}  & 0.34 & $+1.24$ & +0.3 \\
\texttt{GIST-small-Embedding-v0}  & 0.77 & $+1.03$ & +0.3 \\
\texttt{all-MiniLM-L6-v2}  & 0.19 & $+0.89$ & +0.2 \\
\texttt{paraphrase-multilingual-Mi..}  & 0.60 & $+0.85$ & +0.2 \\
\texttt{granite-embedding-30m-engl..}  & 0.69 & $+0.75$ & +0.2 \\
\texttt{GIST-Embedding-v0}  & 0.74 & $+0.66$ & +0.2 \\
\texttt{granite-embedding-125m-eng..}  & 0.82 & $+0.56$ & +0.2 \\
\texttt{potion-base-8M}  & 0.23 & $+0.42$ & +0.1 \\
\texttt{e5-small-v2}  & 0.85 & $+0.16$ & +0.0 \\
\texttt{jina-embeddings-v2-base-en}  & 0.80 & $-0.22$ & -0.1 \\
\texttt{USER-base}  & 0.50 & $-0.28$ & -0.1 \\
\texttt{Squirtle}  & 0.60 & $-0.60$ & -0.2 \\
\texttt{granite-embedding-278m-mul..}  & 0.69 & $-0.87$ & -0.2 \\
\texttt{nomic-embed-text-v1}  & 0.59 & $-0.90$ & -0.2 \\
\texttt{Wartortle}  & 0.62 & $-0.93$ & -0.2 \\
\texttt{e5-small}  & 0.86 & $-1.24$ & -0.3 \\
\texttt{granite-embedding-107m-mul..}  & 0.71 & $-1.51$ & -0.4 \\
\texttt{sentence-t5-base}  & 0.84 & $-2.12$ & -0.6 \\
\texttt{sentence-t5-large}  & 0.81 & $-2.53$ & -0.7 \\
\midrule
\multicolumn{4}{l}{$\spadesuit$ Snowflake family. Family total $297.2$ = 79.7\% of cumulative $\sigma_{\text{cell}}$.} \\
\bottomrule
\end{tabular}
\end{table}

\noindent The five Snowflake-arctic-embed models contribute roughly $80\%$ of the cumulative R2 retrieval $\sigma_{\text{cell}}$ on the $38$-model panel. This concentration is the underlying reason that retrieval is treated as a family-sensitive secondary finding throughout the body, and why we report Snowflake-free numbers in Table~\ref{tab:38model}.

\section{Per-Model Effect Distribution}\label{app:halfviolin}

\begin{figure}[h]
\centering
\includegraphics[width=0.85\textwidth, angle=270, origin=c]{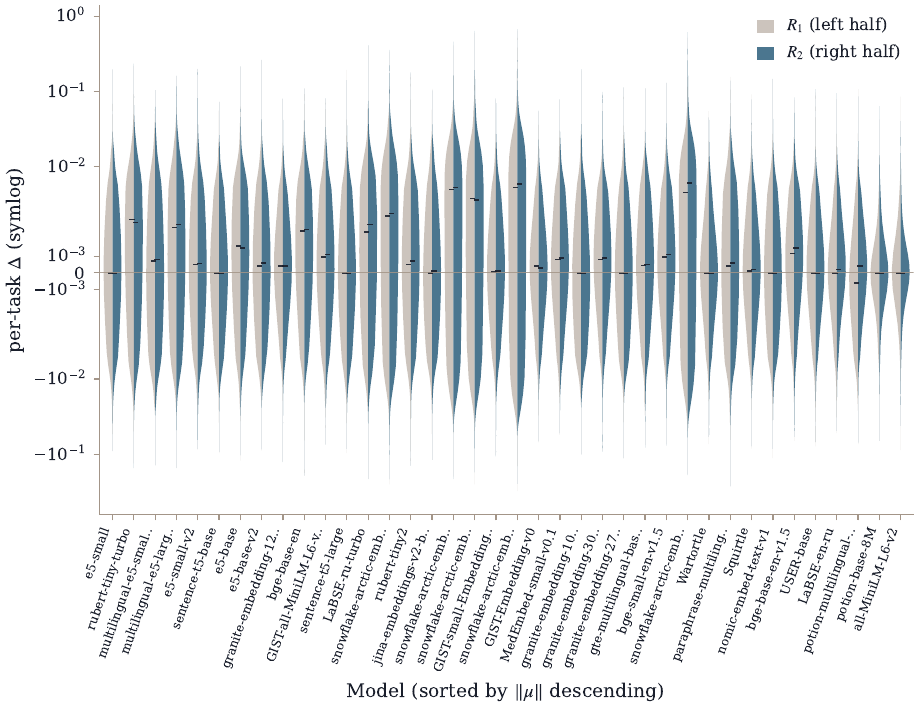}
\caption{Comparison of two renormalization methods across $38$ models, sorted by mean vector norm. Each pair of half-violins shows the per-task score-difference distribution under R1 vs R2 for one model. The median (red bar), quartiles, and adjacent values are marked. The y-axis is logarithm-scaled to better show distributional shape. Models from the same organization share a colour; different organizations use different colours.}
\label{fig:halfviolin}
\end{figure}

\section{$\Vert\mu\Vert$ vs Per-Model Mean $\Delta$, Both Methods}\label{app:norm_vs_delta}

\begin{figure}[h]
\centering
\includegraphics[width=0.7\textwidth]{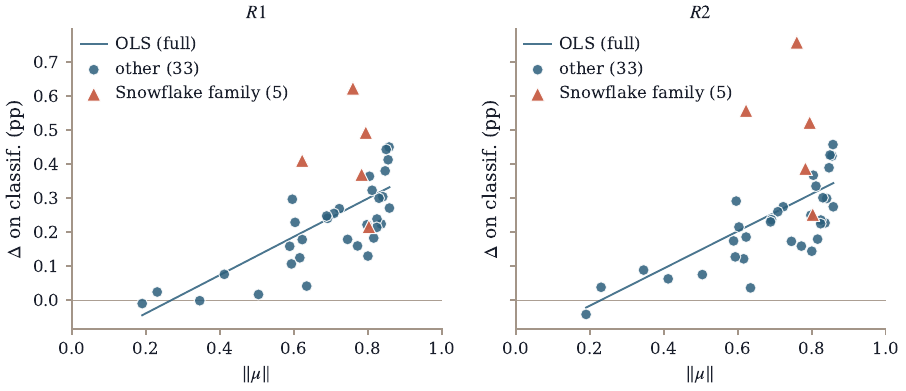}
\caption{Per-model mean task-score change $\Delta$ (log-scaled) under R1 (left) and R2 (right) versus $\Vert\mu\Vert$. Error bars use the per-model dispersion derived from per-task variability. Fitted lines are descriptive only; the per-method correlations $r$ are reported in the body.}
\label{fig:norm_vs_delta_both}
\end{figure}

\begin{figure}[h]
\centering
\includegraphics[width=0.7\textwidth]{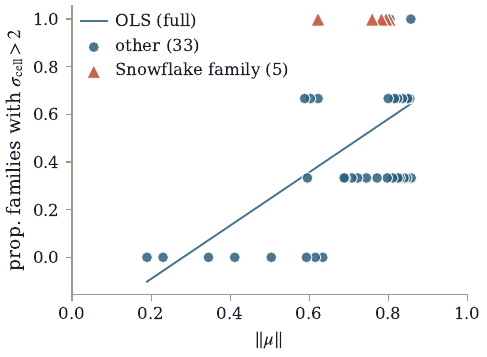}
\caption{Proportion of tasks with $|\sigma_{\text{cell}}|>2$ improvement under R2 versus $\Vert\mu\Vert$ on the $38$-model panel. The body's Figure~\ref{fig:norm_vs_t} reports the same data using paired $t$ on the y-axis; this version uses the secondary $\sigma_{\text{cell}}$ metric (see Appendix~\ref{app:sigma_cell}).}
\label{fig:norm_vs_sigma}
\end{figure}

\section{Per-Model Significance Profile under R2}\label{app:model_perf}

\begin{table}[p]
\centering
\caption{Per-model proportion of tasks where R2 produces a large effect, measured as per-model-internal $|z|>2$ (\,$\Delta$ deviates from the model's own mean $\Delta$ by more than $2$ within-model standard deviations). This is a model-internal effect-size, not the per-family $\sigma_{\text{cell}}$ used elsewhere; computed directly from per-task $\Delta$ in \texttt{RESULTS\_38MODEL.json}. Models sorted by improvement proportion.}
\label{tab:model_performance}
\small
\begin{tabular}{lrrr}
\toprule
Model & $\Vert\mu\Vert$ & $>2\sigma_\Delta$ (\%) & $<-2\sigma_\Delta$ (\%) \\
\midrule
\texttt{snowflake-arctic-embed-xs} & 0.80 & 5 & 0 \\
\texttt{snowflake-arctic-embed-m} & 0.76 & 5 & 0 \\
\texttt{snowflake-arctic-embed-m-v..} & 0.62 & 5 & 0 \\
\texttt{snowflake-arctic-embed-s} & 0.78 & 5 & 0 \\
\texttt{snowflake-arctic-embed-m-l..} & 0.79 & 5 & 0 \\
\texttt{multilingual-e5-large-inst..} & 0.85 & 4 & 1 \\
\texttt{bge-base-en} & 0.82 & 4 & 2 \\
\texttt{Wartortle} & 0.62 & 4 & 3 \\
\texttt{sentence-t5-large} & 0.81 & 3 & 2 \\
\texttt{rubert-tiny-turbo} & 0.86 & 3 & 2 \\
\texttt{bge-base-en-v1.5} & 0.59 & 3 & 2 \\
\texttt{bge-small-en-v1.5} & 0.62 & 3 & 3 \\
\texttt{Squirtle} & 0.60 & 3 & 2 \\
\texttt{multilingual-e5-small} & 0.85 & 3 & 1 \\
\texttt{LaBSE-ru-turbo} & 0.80 & 3 & 1 \\
\texttt{GIST-small-Embedding-v0} & 0.77 & 3 & 2 \\
\texttt{e5-base} & 0.84 & 3 & 1 \\
\texttt{potion-multilingual-128M} & 0.34 & 3 & 2 \\
\texttt{GIST-all-MiniLM-L6-v2} & 0.82 & 3 & 1 \\
\texttt{sentence-t5-base} & 0.84 & 3 & 2 \\
\texttt{granite-embedding-125m-eng..} & 0.82 & 3 & 2 \\
\texttt{paraphrase-multilingual-Mi..} & 0.60 & 3 & 2 \\
\texttt{GIST-Embedding-v0} & 0.74 & 3 & 2 \\
\texttt{all-MiniLM-L6-v2} & 0.19 & 3 & 3 \\
\texttt{jina-embeddings-v2-base-en} & 0.80 & 3 & 2 \\
\texttt{nomic-embed-text-v1} & 0.59 & 3 & 2 \\
\texttt{MedEmbed-small-v0.1} & 0.72 & 3 & 2 \\
\texttt{e5-base-v2} & 0.83 & 3 & 1 \\
\texttt{e5-small-v2} & 0.85 & 3 & 2 \\
\texttt{LaBSE-en-ru} & 0.41 & 2 & 3 \\
\texttt{granite-embedding-278m-mul..} & 0.69 & 2 & 1 \\
\texttt{gte-multilingual-base} & 0.63 & 2 & 3 \\
\texttt{granite-embedding-30m-engl..} & 0.69 & 2 & 2 \\
\texttt{granite-embedding-107m-mul..} & 0.71 & 2 & 0 \\
\texttt{e5-small} & 0.86 & 2 & 3 \\
\texttt{potion-base-8M} & 0.23 & 2 & 2 \\
\texttt{USER-base} & 0.50 & 2 & 3 \\
\texttt{rubert-tiny2} & 0.80 & 2 & 2 \\
\bottomrule
\end{tabular}
\end{table}

\noindent Each row reports, for one of the $28$ audited models with sufficient per-task $\sigma$ data, the proportion of tasks for which R2 yields a significant improvement ($>2\sigma_{\text{cell}}$) and the proportion for which it yields a significant degradation. Highest-$\Vert\mu\Vert$ Snowflake-family models top the list; the $\Vert\mu\Vert\le 0.3$ tail (\texttt{all-MiniLM-L6-v2}, \texttt{potion-base-8M}) is at the bottom.

\section{Per-Task Significance Profile under R2}\label{app:task_impact}

\begin{table}[p]
\centering
\caption{Per-task proportion of the $38$ audited models for which R2 produces a large improvement or degradation, using the same per-model-internal $|z|>2$ metric as Table~\ref{tab:model_performance}. Tasks with no models meeting the threshold are omitted.}
\label{tab:task_impact}
\small
\begin{tabular}{llcc}
\toprule
Task & Type & $>2\sigma_\Delta$ (\%) & $<-2\sigma_\Delta$ (\%) \\
\midrule
STS17 & Other & 50 & 0 \\
ItaCaseholdClassificatio & Classification & 45 & 3 \\
CUADAffiliateLicenseLice & Classification & 42 & 11 \\
MTOPIntentClassification & Classification & 42 & 3 \\
WinoGrande & Retrieval & 42 & 13 \\
SyntecRetrieval & Retrieval & 39 & 0 \\
CUADUnlimitedAllYouCanEa & Classification & 29 & 11 \\
ChemNQRetrieval & Retrieval & 29 & 8 \\
LegalReasoningCausalityL & Classification & 29 & 13 \\
ChemHotpotQARetrieval & Retrieval & 26 & 32 \\
SynPerChatbotConvSAToneU & Classification & 26 & 0 \\
FQuADRetrieval & Retrieval & 26 & 0 \\
WikipediaSpecialtiesInCh & Other & 26 & 13 \\
Touche2020Retrieval.v3 & Retrieval & 24 & 0 \\
NLPTwitterAnalysisClassi & Classification & 24 & 0 \\
WikipediaChemistryTopics & Classification & 24 & 26 \\
SyntecReranking & Other & 21 & 3 \\
HotpotQAHardNegatives & Retrieval & 21 & 0 \\
LEMBWikimQARetrieval & Retrieval & 21 & 0 \\
BigPatentClustering.v2 & Other & 21 & 3 \\
NQHardNegatives & Retrieval & 18 & 3 \\
SKQuadRetrieval & Retrieval & 18 & 11 \\
AlloprofRetrieval & Retrieval & 18 & 0 \\
ContractNLIInclusionOfVe & Classification & 18 & 0 \\
CUADExclusivityLegalBenc & Classification & 18 & 0 \\
GermanDPR & Retrieval & 18 & 3 \\
TextualismToolDictionari & Classification & 18 & 53 \\
FEVERHardNegatives & Retrieval & 18 & 18 \\
EstQA & Retrieval & 18 & 0 \\
CanadaTaxCourtOutcomesLe & Classification & 16 & 5 \\
KlueMrcDomainClustering & Other & 16 & 13 \\
SlovakSumRetrieval & Retrieval & 16 & 0 \\
MedicalQARetrieval & Retrieval & 16 & 0 \\
SpanishNewsClusteringP2P & Other & 16 & 29 \\
ContractNLIPermissibleCo & Classification & 16 & 13 \\
SyntheticText2SQL & Retrieval & 16 & 11 \\
BuiltBenchClusteringP2P & Other & 16 & 26 \\
LitSearchRetrieval & Retrieval & 16 & 0 \\
Banking77Classification & Classification & 16 & 3 \\
HotpotQA-PLHardNegatives & Retrieval & 16 & 3 \\
CUADWarrantyDurationLega & Classification & 13 & 0 \\
CUADSourceCodeEscrowLega & Classification & 13 & 11 \\
SCDBPTrainingLegalBenchC & Classification & 13 & 5 \\
SynPerChatbotToneUserCla & Classification & 13 & 0 \\
CQADupstackGamingRetriev & Retrieval & 13 & 0 \\
FeedbackQARetrieval & Retrieval & 13 & 0 \\
DBPediaHardNegatives & Retrieval & 13 & 0 \\
TV2Nordretrieval & Retrieval & 13 & 3 \\
KlueYnatMrcCategoryClust & Other & 13 & 16 \\
BSARDRetrieval & Retrieval & 13 & 8 \\
SynPerChatbotRAGToneUser & Classification & 13 & 0 \\
CUADThirdPartyBeneficiar & Classification & 13 & 11 \\
SCDDAccountabilityLegalB & Classification & 13 & 0 \\
CQADupstackPhysicsRetrie & Retrieval & 13 & 0 \\
ContractNLISurvivalOfObl & Classification & 13 & 5 \\
GermanGovServiceRetrieva & Retrieval & 13 & 0 \\
JCrewBlockerLegalBenchCl & Classification & 13 & 0 \\
LegalBenchCorporateLobby & Retrieval & 13 & 0 \\
CQADupstackGisRetrieval & Retrieval & 11 & 0 \\
MSMARCOHardNegatives & Retrieval & 11 & 0 \\
\bottomrule
\end{tabular}
\end{table}

\noindent Each row reports, for one MMTEB task, the proportion of audited models for which R2 yields a significant improvement or degradation on that task. Retrieval and clustering tasks dominate the top of the list; a small tail of pair-classification, summarization, and reranking tasks shows mixed effects.

\section{Large-Effect Tasks: Per-Model Breakdown}\label{app:model_perf2}

\begin{table*}[ht]
\centering
\caption{Per-model large-effect tasks under R2 (filter: $|\Delta|>0.1$ score units AND $|\delta|>2\%$ relative change). Computed from \texttt{RESULTS\_38MODEL.json}. Only models with at least one task meeting the filter are shown.}
\label{tab:model_performance2}
\scriptsize
\begin{tabular}{lrrrrrrrr}
\toprule
Model & $\#\uparrow$ & $\bar{\Delta}_\uparrow$ & $\Delta_{\max,\uparrow}$ & $\Delta_{\min,\uparrow}$ & $\#\downarrow$ & $\bar{\Delta}_\downarrow$ & $\Delta_{\max,\downarrow}$ & $\Delta_{\min,\downarrow}$ \\
\midrule
\texttt{snowflake-arctic-embed-m} & 59 & 0.2346 & 0.5849 & 0.1019 & 3 & -0.1462 & -0.1264 & -0.1589 \\
\texttt{snowflake-arctic-embed-m-v..} & 53 & 0.1908 & 0.5338 & 0.1011 & 2 & -0.1058 & -0.1034 & -0.1081 \\
\texttt{snowflake-arctic-embed-s} & 43 & 0.1625 & 0.5559 & 0.1021 & 0 & -- & -- & -- \\
\texttt{snowflake-arctic-embed-m-l..} & 39 & 0.1905 & 0.3992 & 0.1022 & 2 & -0.1341 & -0.1149 & -0.1532 \\
\texttt{snowflake-arctic-embed-xs} & 29 & 0.1536 & 0.3092 & 0.1011 & 1 & -0.1682 & -0.1682 & -0.1682 \\
\texttt{multilingual-e5-large-inst..} & 4 & 0.1192 & 0.1407 & 0.1082 & 1 & -0.1308 & -0.1308 & -0.1308 \\
\texttt{LaBSE-ru-turbo} & 3 & 0.1982 & 0.3586 & 0.1150 & 1 & -0.1589 & -0.1589 & -0.1589 \\
\texttt{rubert-tiny-turbo} & 3 & 0.1573 & 0.2028 & 0.1102 & 1 & -0.1215 & -0.1215 & -0.1215 \\
\texttt{rubert-tiny2} & 2 & 0.1362 & 0.1553 & 0.1171 & 1 & -0.1192 & -0.1192 & -0.1192 \\
\texttt{e5-base} & 2 & 0.1489 & 0.1871 & 0.1108 & 0 & -- & -- & -- \\
\texttt{e5-base-v2} & 2 & 0.1740 & 0.2289 & 0.1191 & 0 & -- & -- & -- \\
\texttt{nomic-embed-text-v1} & 2 & 0.1148 & 0.1271 & 0.1025 & 0 & -- & -- & -- \\
\texttt{paraphrase-multilingual-Mi..} & 2 & 0.1034 & 0.1040 & 0.1028 & 1 & -0.2316 & -0.2316 & -0.2316 \\
\texttt{bge-small-en-v1.5} & 1 & 0.1130 & 0.1130 & 0.1130 & 0 & -- & -- & -- \\
\texttt{granite-embedding-107m-mul..} & 1 & 0.1723 & 0.1723 & 0.1723 & 0 & -- & -- & -- \\
\texttt{e5-small} & 1 & 0.1709 & 0.1709 & 0.1709 & 0 & -- & -- & -- \\
\texttt{e5-small-v2} & 1 & 0.1729 & 0.1729 & 0.1729 & 0 & -- & -- & -- \\
\texttt{sentence-t5-large} & 1 & 0.1227 & 0.1227 & 0.1227 & 0 & -- & -- & -- \\
\texttt{bge-base-en} & 0 & -- & -- & -- & 1 & -0.1669 & -0.1669 & -0.1669 \\
\texttt{granite-embedding-125m-eng..} & 0 & -- & -- & -- & 1 & -0.1101 & -0.1101 & -0.1101 \\
\texttt{multilingual-e5-small} & 0 & -- & -- & -- & 1 & -0.1215 & -0.1215 & -0.1215 \\
\texttt{jina-embeddings-v2-base-en} & 0 & -- & -- & -- & 1 & -0.1028 & -0.1028 & -0.1028 \\
\bottomrule
\end{tabular}
\end{table*}

\noindent Filter: $|\Delta|>0.1$ score units AND relative change $|\delta|>2\%$. The first numeric column is the count of tasks with significant improvement; the columns to its right are the average, max, and min improvement deltas. The columns starting with $\#\downarrow$ are the analogous degradation statistics. Models not listed had no tasks meeting the filter.

\section{Large-Effect Tasks: Per-Task Breakdown}\label{app:task_impact2}

\begin{table*}[ht]
\centering
\caption{Per-task large-effect breakdown under R2 (filter: $|\Delta|>0.1$ score units AND $|\delta|>2\%$ relative change). Computed from \texttt{RESULTS\_38MODEL.json}.}
\label{tab:task_impact2}
\footnotesize
\begin{tabular}{lrrrrrrrrr}
\toprule
Task & $N$ & $\#\uparrow$ & $\bar{\Delta}_\uparrow$ & $\Delta_{\max,\uparrow}$ & $\Delta_{\min,\uparrow}$ & $\#\downarrow$ & $\bar{\Delta}_\downarrow$ & $\Delta_{\max,\downarrow}$ & $\Delta_{\min,\downarrow}$ \\
\midrule
WinoGrande & 10 & 10 & 0.2121 & 0.5559 & 0.1150 & 0 & -- & -- & -- \\
NQHardNegatives & 5 & 5 & 0.2532 & 0.3002 & 0.1909 & 0 & -- & -- & -- \\
CQADupstackTexRetrieval & 5 & 5 & 0.1320 & 0.1772 & 0.1059 & 0 & -- & -- & -- \\
MSMARCOHardNegatives & 5 & 5 & 0.1435 & 0.1830 & 0.1223 & 0 & -- & -- & -- \\
ChemNQRetrieval & 5 & 5 & 0.2360 & 0.2884 & 0.1785 & 0 & -- & -- & -- \\
CQADupstackGamingRetriev & 5 & 5 & 0.1963 & 0.3450 & 0.1295 & 0 & -- & -- & -- \\
FeedbackQARetrieval & 5 & 5 & 0.2353 & 0.3549 & 0.1270 & 0 & -- & -- & -- \\
HotpotQAHardNegatives & 5 & 5 & 0.3459 & 0.4562 & 0.2573 & 0 & -- & -- & -- \\
FiQA2018 & 5 & 5 & 0.1620 & 0.1944 & 0.1315 & 0 & -- & -- & -- \\
ChemHotpotQARetrieval & 6 & 5 & 0.4234 & 0.5849 & 0.2728 & 1 & -0.1192 & -0.1192 & -0.1192 \\
CQADupstackEnglishRetrie & 5 & 5 & 0.1562 & 0.2108 & 0.1240 & 0 & -- & -- & -- \\
CQADupstackProgrammersRe & 5 & 5 & 0.1544 & 0.2198 & 0.1321 & 0 & -- & -- & -- \\
SyntheticText2SQL & 5 & 5 & 0.1834 & 0.2605 & 0.1254 & 0 & -- & -- & -- \\
CQADupstackPhysicsRetrie & 5 & 5 & 0.1772 & 0.2824 & 0.1180 & 0 & -- & -- & -- \\
Touche2020Retrieval.v3 & 5 & 5 & 0.2245 & 0.2698 & 0.1903 & 0 & -- & -- & -- \\
CQADupstackWordpressRetr & 5 & 5 & 0.1491 & 0.2460 & 0.1093 & 0 & -- & -- & -- \\
CQADupstackRetrieval & 5 & 5 & 0.1508 & 0.2450 & 0.1099 & 0 & -- & -- & -- \\
LitSearchRetrieval & 5 & 5 & 0.3329 & 0.3992 & 0.2690 & 0 & -- & -- & -- \\
CQADupstackUnixRetrieval & 5 & 5 & 0.1448 & 0.2557 & 0.1080 & 0 & -- & -- & -- \\
HotpotQA-PLHardNegatives & 5 & 5 & 0.1436 & 0.1743 & 0.1011 & 0 & -- & -- & -- \\
WikipediaSpecialtiesInCh & 6 & 5 & 0.1525 & 0.2289 & 0.1130 & 1 & -0.1669 & -0.1669 & -0.1669 \\
SyntecRetrieval & 5 & 5 & 0.2100 & 0.3544 & 0.1046 & 0 & -- & -- & -- \\
FEVERHardNegatives & 5 & 5 & 0.3113 & 0.5009 & 0.1407 & 0 & -- & -- & -- \\
CQADupstackGisRetrieval & 4 & 4 & 0.1655 & 0.2725 & 0.1039 & 0 & -- & -- & -- \\
MedicalQARetrieval & 4 & 4 & 0.2924 & 0.3584 & 0.1618 & 0 & -- & -- & -- \\
MLQuestions & 4 & 4 & 0.1694 & 0.2086 & 0.1163 & 0 & -- & -- & -- \\
NFCorpus & 4 & 4 & 0.2106 & 0.2569 & 0.1299 & 0 & -- & -- & -- \\
DBPediaHardNegatives & 4 & 4 & 0.1995 & 0.2422 & 0.1471 & 0 & -- & -- & -- \\
SCIDOCS & 4 & 4 & 0.1327 & 0.1471 & 0.1154 & 0 & -- & -- & -- \\
CUADThirdPartyBeneficiar & 4 & 4 & 0.1434 & 0.1618 & 0.1029 & 0 & -- & -- & -- \\
SciFact & 4 & 4 & 0.3423 & 0.5767 & 0.1504 & 0 & -- & -- & -- \\
CQADupstackMathematicaRe & 4 & 4 & 0.1384 & 0.1854 & 0.1154 & 0 & -- & -- & -- \\
LEMBWikimQARetrieval & 4 & 4 & 0.1961 & 0.3208 & 0.1479 & 0 & -- & -- & -- \\
CQADupstackStatsRetrieva & 4 & 4 & 0.1485 & 0.1972 & 0.1244 & 0 & -- & -- & -- \\
LegalBenchCorporateLobby & 4 & 4 & 0.2605 & 0.5331 & 0.1023 & 0 & -- & -- & -- \\
SKQuadRetrieval & 3 & 3 & 0.1651 & 0.1768 & 0.1424 & 0 & -- & -- & -- \\
SweFaqRetrieval & 3 & 3 & 0.1139 & 0.1277 & 0.1011 & 0 & -- & -- & -- \\
SlovakSumRetrieval & 3 & 3 & 0.1336 & 0.1577 & 0.1019 & 0 & -- & -- & -- \\
TV2Nordretrieval & 3 & 3 & 0.1474 & 0.1705 & 0.1134 & 0 & -- & -- & -- \\
BSARDRetrieval & 3 & 3 & 0.1276 & 0.1441 & 0.1036 & 0 & -- & -- & -- \\
SwednRetrieval & 3 & 3 & 0.1342 & 0.1522 & 0.1202 & 0 & -- & -- & -- \\
BuiltBenchRetrieval & 3 & 3 & 0.2456 & 0.4602 & 0.1104 & 0 & -- & -- & -- \\
CQADupstackWebmastersRet & 3 & 3 & 0.1689 & 0.2543 & 0.1090 & 0 & -- & -- & -- \\
GermanDPR & 3 & 3 & 0.1210 & 0.1442 & 0.1093 & 0 & -- & -- & -- \\
FQuADRetrieval & 3 & 3 & 0.2077 & 0.2391 & 0.1705 & 0 & -- & -- & -- \\
CQADupstackAndroidRetrie & 3 & 3 & 0.1811 & 0.2940 & 0.1022 & 0 & -- & -- & -- \\
ClimateFEVERHardNegative & 3 & 3 & 0.1654 & 0.2016 & 0.1128 & 0 & -- & -- & -- \\
PubChemWikiPairClassific & 3 & 3 & 0.1681 & 0.2348 & 0.1213 & 0 & -- & -- & -- \\
GermanGovServiceRetrieva & 3 & 3 & 0.1582 & 0.2028 & 0.1040 & 0 & -- & -- & -- \\
WikipediaRetrievalMultil & 2 & 2 & 0.1333 & 0.1490 & 0.1177 & 0 & -- & -- & -- \\
ZacLegalTextRetrieval & 2 & 2 & 0.1273 & 0.1464 & 0.1082 & 0 & -- & -- & -- \\
SyntecReranking & 2 & 2 & 0.1537 & 0.2028 & 0.1045 & 0 & -- & -- & -- \\
NarrativeQARetrieval & 2 & 2 & 0.1137 & 0.1138 & 0.1136 & 0 & -- & -- & -- \\
LEMBNarrativeQARetrieval & 2 & 2 & 0.1135 & 0.1138 & 0.1132 & 0 & -- & -- & -- \\
GermanQuAD-Retrieval & 2 & 2 & 0.1883 & 0.2039 & 0.1727 & 0 & -- & -- & -- \\
CUADJointIPOwnershipLega & 2 & 2 & 0.1849 & 0.2448 & 0.1250 & 0 & -- & -- & -- \\
SciFact-NL & 2 & 2 & 0.1417 & 0.1558 & 0.1277 & 0 & -- & -- & -- \\
WikipediaChemistryTopics & 2 & 2 & 0.1074 & 0.1108 & 0.1040 & 0 & -- & -- & -- \\
LegalBenchConsumerContra & 1 & 1 & 0.1823 & 0.1823 & 0.1823 & 0 & -- & -- & -- \\
LEMBQMSumRetrieval & 1 & 1 & 0.1030 & 0.1030 & 0.1030 & 0 & -- & -- & -- \\
AlloprofRetrieval & 1 & 1 & 0.1379 & 0.1379 & 0.1379 & 0 & -- & -- & -- \\
MLQARetrieval & 1 & 1 & 0.1135 & 0.1135 & 0.1135 & 0 & -- & -- & -- \\
SynPerChatbotConvSAHappi & 1 & 1 & 0.1025 & 0.1025 & 0.1025 & 0 & -- & -- & -- \\
ToxicChatClassification & 1 & 1 & 0.1040 & 0.1040 & 0.1040 & 0 & -- & -- & -- \\
TurHistQuadRetrieval & 1 & 1 & 0.1021 & 0.1021 & 0.1021 & 0 & -- & -- & -- \\
CUADNoSolicitOfCustomers & 1 & 1 & 0.1071 & 0.1071 & 0.1071 & 0 & -- & -- & -- \\
CodeSearchNetRetrieval & 1 & 1 & 0.1441 & 0.1441 & 0.1441 & 0 & -- & -- & -- \\
BelebeleRetrieval & 1 & 1 & 0.1057 & 0.1057 & 0.1057 & 0 & -- & -- & -- \\
CosQA & 1 & 1 & 0.1405 & 0.1405 & 0.1405 & 0 & -- & -- & -- \\
LegalSummarization & 1 & 1 & 0.1035 & 0.1035 & 0.1035 & 0 & -- & -- & -- \\
TextualismToolDictionari & 8 & 1 & 0.1028 & 0.1028 & 0.1028 & 7 & -0.1375 & -0.1028 & -0.1682 \\
CUADNoticePeriodToTermin & 1 & 1 & 0.1171 & 0.1171 & 0.1171 & 0 & -- & -- & -- \\
STSES & 1 & 1 & 0.1215 & 0.1215 & 0.1215 & 0 & -- & -- & -- \\
TempReasonL2Fact & 1 & 1 & 0.1150 & 0.1150 & 0.1150 & 0 & -- & -- & -- \\
SummEvalSummarization.v2 & 1 & 1 & 0.1121 & 0.1121 & 0.1121 & 0 & -- & -- & -- \\
IndicCrosslingualSTS & 2 & 1 & 0.1209 & 0.1209 & 0.1209 & 1 & -0.1101 & -0.1101 & -0.1101 \\
RARbCode & 1 & 1 & 0.1577 & 0.1577 & 0.1577 & 0 & -- & -- & -- \\
EstQA & 1 & 1 & 0.1311 & 0.1311 & 0.1311 & 0 & -- & -- & -- \\
ContractNLIPermissiblePo & 3 & 0 & -- & -- & -- & 3 & -0.1382 & -0.1081 & -0.1532 \\
ContractNLIPermissibleCo & 3 & 0 & -- & -- & -- & 3 & -0.1149 & -0.1034 & -0.1264 \\
\bottomrule
\end{tabular}
\end{table*}

\noindent Same filter as Appendix~\ref{app:model_perf2}, but indexed by task rather than model. \texttt{WinoGrande} is uniformly improved across all $10$ models for which it is in the test grid; \texttt{ChemHotpotQARetrieval}, the \texttt{CQADupstack*} family, and \texttt{HotpotQAHardNegatives} also show broad gains.

\section{Per-Model 9-Method Task Counts}\label{app:9method_tasks}

\begin{table}[ht]
\centering
\caption{Per-model task completion across the $9$-method ladder. The body of Table~2 reports $\Delta$ on the per-model intersection (right-most column).}
\label{tab:9method_task_counts}
\scriptsize
\setlength{\tabcolsep}{4pt}
\begin{tabular}{lcrrrrrrrrrr}
\toprule
Model & $\Vert\mu\Vert$ & control & R1 & R2 & ABTT-1 & ABTT-2 & ABTT-3 & mc & rand & whi & intersection \\
\midrule
\texttt{multilingual-e5-large-inst..} & 0.85 & 603 & 725 & 516 & 649 & 572 & 559 & 615 & 650 & 473 & 462 \\
\texttt{snowflake-arctic-embed-m} & 0.76 & 555 & 569 & 589 & 593 & 561 & 555 & 579 & 553 & 444 & 333 \\
\texttt{bge-base-en-v1.5} & 0.59 & 667 & 648 & 649 & 582 & 593 & 649 & 557 & 474 & 447 & 446 \\
\texttt{nomic-embed-text-v1} & 0.59 & 608 & 582 & 596 & 596 & 569 & 491 & 564 & 569 & 402 & 402 \\
\texttt{all-MiniLM-L6-v2} & 0.19 & 452 & 597 & 623 & 600 & 608 & 601 & 597 & 579 & 603 & 443 \\
\bottomrule
\end{tabular}
\end{table}

\noindent The right-most column is the per-model intersection used as the basis for the $\Delta$ values in Table~\ref{tab:9method}. Per-method counts vary because some methods experience occasional task-level failures (timeouts, OOMs, or numerical errors specific to the post-processing step); in particular, the \texttt{whiten} column has the smallest task count for every model because full-rank PCA on the largest tasks intermittently exceeded our memory budget.

\section{MMTEB Task Filtering: Full $192$-Task List}\label{app:filtered}

The $5$-model nine-method ablation uses MMTEB v$1.38.18$ ($929$ tasks total) and excludes $192$ tasks falling into three disjoint categories:
\begin{enumerate}
    \item \emph{Failures} ($115$): tasks that consistently crash or produce invalid outputs across multiple runs.
    \item \emph{Timeouts} ($28$): tasks that exceed the wall-clock budget (default $1$\,h, sometimes extended to $3$\,h) even after retries.
    \item \emph{Unsupported modality} ($49$): image-based or multimodal benchmarks incompatible with text-only embedding models.
\end{enumerate}
The $38$-model R1/R2 audit was run on an earlier MMTEB snapshot; per-model R2 task counts are $609$ to $624$ after the same three filtering categories. Per-study task-tracker metadata accompanies the submission.

\begin{table}[h]
\centering
\caption{Tasks filtered from MMTEB evaluation (static metadata, identical for all runs). Three disjoint categories: reproducible failures, wall-clock timeouts, and unsupported (image/multimodal) modalities. This list is for the MMTEB v1.38.18 snapshot used by the $5$-model nine-method ablation; the $38$-model audit used an earlier snapshot with a similar but not identical filter set (per-study task-tracker JSONs accompany the submission).}
\label{tab:filtered_tasks}
\scriptsize
\setlength{\tabcolsep}{3pt}
\begin{tabularx}{\textwidth}{l c Y}
\toprule
Category & Count & Tasks\\
\midrule
Failures & 115 & VisualSTS17Eng, VisualSTS17Multilingual, VisualSTS-b-Eng, VisualSTS-b-Multilingual, CodeEditSearchRetrieval, CodeRAGProgrammingSolutions, CodeRAGOnlineTutorials, CodeRAGLibraryDocumentationSolutions, DanFeverRetrieval, TwitterHjerneRetrieval, GreekCivicsQA, BrightRetrieval, BrightLongRetrieval, ClimateFEVER.v2, HagridRetrieval, LEMBNeedleRetrieval, LEMBPasskeyRetrieval, LEMBSummScreenFDRetrieval, MSMARCO, NanoArguAnaRetrieval, NanoClimateFeverRetrieval, NanoDBPediaRetrieval, NanoFEVERRetrieval, NanoFiQA2018Retrieval, NanoHotpotQARetrieval, NanoMSMARCORetrieval, NanoNFCorpusRetrieval, NanoNQRetrieval, NanoQuoraRetrieval, NanoSCIDOCSRetrieval, NanoSciFactRetrieval, NanoTouche2020Retrieval, TopiOCQA, TopiOCQAHardNegatives, MSMARCO-Fa, JaQuADRetrieval, Ko-StrategyQA, CUREv1, MIRACLRetrievalHardNegatives, NeuCLIR2022Retrieval, NeuCLIR2023Retrieval, WebFAQRetrieval, XQuADRetrieval, mMARCO-NL, Touche2020-NL, SNLRetrieval, SpanishPassageRetrievalS2P, SpanishPassageRetrievalS2S, VieQuADRetrieval, T2Retrieval, MMarcoRetrieval, DuRetrieval, CovidRetrieval, CmedqaRetrieval, EcomRetrieval, MedicalRetrieval, VideoRetrieval, WebQAT2TRetrieval, DKHateClassification, FinancialPhrasebankClassification, TweetTopicSingleClassification, DigikalamagClassification, FrenchBookReviews, HindiDiscourseClassification, SentimentAnalysisHindi, IndonesianIdClickbaitClassification, KannadaNewsClassification, KLUE-TC, KorFin, AfriSentiLangClassification, TurkicClassification, MyanmarNews, HateSpeechPortugueseClassification, SanskritShlokasClassification, SinhalaNewsClassification, SinhalaNewsSourceClassification, SpanishNewsClassification, SiswatiNewsClassification, SwahiliNewsClassification, UrduRomanSentimentClassification, IsiZuluNewsClassification, BibleNLPBitextMining, FloresBitextMining, IWSLT2017BitextMining, LinceMTBitextMining, NollySentiBitextMining, NusaTranslationBitextMining, NusaXBitextMining, PhincBitextMining, WebFAQBitextMiningQuestions, WebFAQBitextMiningQAs, DigikalamagClustering, NLPTwitterAnalysisClustering, SNLHierarchicalClusteringP2P, SNLHierarchicalClusteringS2S, SwednClusteringP2P, SwednClusteringS2S, PubChemSMILESPC, indonli, KLUE-NLI, TERRa, Ocnli, Cmnli, WebLINXCandidatesReranking, MIRACLReranking, T2Reranking, MMarcoReranking, CPUSpeedTask, GPUSpeedTask, FaroeseSTS, JSTS, KLUE-STS, MLSUMClusteringP2P.v2, SIB200ClusteringS2S, WikiClusteringP2P.v2\\
Timeouts & 28 & CodeRAGStackoverflowPosts, MSMARCOv2, NQ, ClimateFEVER-Fa, DBPedia-Fa, HotpotQA-Fa, NQ-Fa, MIRACLRetrieval, MrTidyRetrieval, MultiLongDocRetrieval, ClimateFEVER-NL, DBPedia-NL, FEVER-NL, HotpotQA-NL, NQ-NL, DBPedia-PL, HotpotQA-PL, MSMARCO-PL, NQ-PL, ClimateFEVER, DBPedia, FEVER, HotpotQA, RiaNewsRetrieval, mFollowIRCrossLingualInstructionRetrieval, mFollowIRInstructionRetrieval, MultiEURLEXMultilabelClassification, GerDaLIR\\
Unsupported modality & 49 & CUB200I2IRetrieval, FORBI2IRetrieval, GLDv2I2IRetrieval, METI2IRetrieval, NIGHTSI2IRetrieval, ROxfordEasyI2IRetrieval, ROxfordMediumI2IRetrieval, ROxfordHardI2IRetrieval, RP2kI2IRetrieval, RParisEasyI2IRetrieval, RParisMediumI2IRetrieval, RParisHardI2IRetrieval, SketchyI2IRetrieval, SOPI2IRetrieval, StanfordCarsI2IRetrieval, Birdsnap, Caltech101, CIFAR10, CIFAR100, Country211, DTD, EuroSAT, FER2013, FGVCAircraft, Food101Classification, GTSRB, Imagenet1k, MNIST, OxfordFlowersClassification, OxfordPets, PatchCamelyon, RESISC45, StanfordCars, STL10, SUN397, UCF101, CIFAR10Clustering, CIFAR100Clustering, ImageNetDog15Clustering, ImageNet10Clustering, TinyImageNetClustering, VOC2007, STS12VisualSTS, STS13VisualSTS, STS14VisualSTS, STS15VisualSTS, STS16VisualSTS, STS17MultilingualVisualSTS, STSBenchmarkMultilingualVisualSTS\\
\bottomrule
\end{tabularx}
\end{table}

\section{MMTEB Models Evaluated: Full Panel}\label{app:models}

\begin{table}[ht]
\centering
\caption{MMTEB models evaluated in this study, sorted by official MTEB leaderboard rank (as of arXiv v1). $\Vert\mu\Vert$ values computed from $10^5$ English Wikipedia sentences.}
\label{tab:evaluated_models}
\scriptsize
\setlength{\tabcolsep}{4pt}
\begin{tabularx}{\textwidth}{Y c c c}
\toprule
Model & Size(MB) & $\Vert\mu\Vert$ & MTEB rank\\
\midrule
\href{https://huggingface.co/intfloat/multilingual-e5-large-instruct}{intfloat/multilingual-e5-large-instruct} & 1068 & 0.8491 & 7\\\\
\href{https://huggingface.co/Alibaba-NLP/gte-multilingual-base}{Alibaba-NLP/gte-multilingual-base} & 582 & 0.6341 & 26\\\\
\href{https://huggingface.co/intfloat/multilingual-e5-small}{intfloat/multilingual-e5-small} & 449 & 0.8543 & 38\\\\
\href{https://huggingface.co/ibm-granite/granite-embedding-278m-multilingual}{ibm-granite/granite-embedding-278m-multilingual} & 530 & 0.6882 & 44\\\\
\href{https://huggingface.co/ibm-granite/granite-embedding-107m-multilingual}{ibm-granite/granite-embedding-107m-multilingual} & 204 & 0.7078 & 52\\\\
\href{https://huggingface.co/nomic-ai/nomic-embed-text-v1}{nomic-ai/nomic-embed-text-v1} & 522 & 0.5929 & 57\\\\
\href{https://huggingface.co/intfloat/e5-base-v2}{intfloat/e5-base-v2} & 418 & 0.8297 & 59\\\\
\href{https://huggingface.co/sentence-transformers/paraphrase-multilingual-MiniLM-L12-v2}{sentence-transformers/paraphrase-multilingual-MiniLM-L12-v2} & 449 & 0.6028 & 64\\\\
\href{https://huggingface.co/avsolatorio/GIST-Embedding-v0}{avsolatorio/GIST-Embedding-v0} & 418 & 0.7447 & 67\\\\
\href{https://huggingface.co/BAAI/bge-base-en-v1.5}{BAAI/bge-base-en-v1.5} & 390 & 0.5883 & 75\\\\
\href{https://huggingface.co/avsolatorio/GIST-small-Embedding-v0}{avsolatorio/GIST-small-Embedding-v0} & 127 & 0.7716 & 76\\\\
\href{https://huggingface.co/minishlab/potion-multilingual-128M}{minishlab/potion-multilingual-128M} & 489 & 0.3450 & 79\\\\
\href{https://huggingface.co/intfloat/e5-small-v2}{intfloat/e5-small-v2} & 127 & 0.8464 & 82\\\\
\href{https://huggingface.co/intfloat/e5-base}{intfloat/e5-base} & 418 & 0.8357 & 83\\\\
\href{https://huggingface.co/BAAI/bge-small-en-v1.5}{BAAI/bge-small-en-v1.5} & 127 & 0.6222 & 86\\\\
\href{https://huggingface.co/abhinand/MedEmbed-small-v0.1}{abhinand/MedEmbed-small-v0.1} & 127 & 0.7225 & 91\\\\
\href{https://huggingface.co/ibm-granite/granite-embedding-125m-english}{ibm-granite/granite-embedding-125m-english} & 238 & 0.8242 & 92\\\\
\href{https://huggingface.co/intfloat/e5-small}{intfloat/e5-small} & 127 & 0.8579 & 93\\\\
\href{https://huggingface.co/Snowflake/snowflake-arctic-embed-m-long}{Snowflake/snowflake-arctic-embed-m-long} & 522 & 0.7941 & 98\\\\
\href{https://huggingface.co/avsolatorio/GIST-all-MiniLM-L6-v2}{avsolatorio/GIST-all-MiniLM-L6-v2} & 87 & 0.8154 & 100\\\\
\href{https://huggingface.co/sergeyzh/LaBSE-ru-turbo}{sergeyzh/LaBSE-ru-turbo} & 490 & 0.8039 & 101\\\\
\href{https://huggingface.co/Snowflake/snowflake-arctic-embed-m}{Snowflake/snowflake-arctic-embed-m} & 415 & 0.7592 & 103\\\\
\href{https://huggingface.co/Snowflake/snowflake-arctic-embed-s}{Snowflake/snowflake-arctic-embed-s} & 127 & 0.7826 & 104\\\\
\href{https://huggingface.co/Snowflake/snowflake-arctic-embed-m-v1.5}{Snowflake/snowflake-arctic-embed-m-v1.5} & 415 & 0.6221 & 107\\\\
\href{https://huggingface.co/cointegrated/LaBSE-en-ru}{cointegrated/LaBSE-en-ru} & 492 & 0.4115 & 108\\\\
\href{https://huggingface.co/ibm-granite/granite-embedding-30m-english}{ibm-granite/granite-embedding-30m-english} & 58 & 0.6907 & 110\\\\
\href{https://huggingface.co/sentence-transformers/all-MiniLM-L6-v2}{sentence-transformers/all-MiniLM-L6-v2} & 87 & 0.1895 & 117\\\\
\href{https://huggingface.co/Mihaiii/Wartortle}{Mihaiii/Wartortle} & 66 & 0.6155 & 118\\\\
\href{https://huggingface.co/deepvk/USER-base}{deepvk/USER-base} & 473 & 0.5039 & 120\\\\
\href{https://huggingface.co/Snowflake/snowflake-arctic-embed-xs}{Snowflake/snowflake-arctic-embed-xs} & 86 & 0.8023 & 124\\\\
\href{https://huggingface.co/Mihaiii/Squirtle}{Mihaiii/Squirtle} & 60 & 0.5954 & 128\\\\
\href{https://huggingface.co/sergeyzh/rubert-tiny-turbo}{sergeyzh/rubert-tiny-turbo} & 111 & 0.8570 & 130\\\\
\href{https://huggingface.co/minishlab/potion-base-8M}{minishlab/potion-base-8M} & 29 & 0.2301 & 131\\\\
\href{https://huggingface.co/cointegrated/rubert-tiny2}{cointegrated/rubert-tiny2} & 112 & 0.7997 & 138\\\\
\href{https://huggingface.co/jinaai/jina-embeddings-v2-base-en}{jinaai/jina-embeddings-v2-base-en} & 262 & 0.7972 & 154\\\\
\href{https://huggingface.co/sentence-transformers/sentence-t5-large}{sentence-transformers/sentence-t5-large} & 639 & 0.8110 & 179\\\\
\href{https://huggingface.co/BAAI/bge-base-en}{BAAI/bge-base-en} & 390 & 0.8237 & 189\\\\
\href{https://huggingface.co/sentence-transformers/sentence-t5-base}{sentence-transformers/sentence-t5-base} & 209 & 0.8399 & 192\\\\
\bottomrule
\end{tabularx}
\end{table}

\noindent The $38$-model panel was built by sweeping the public MMTEB leaderboard, then preferentially adding models with $\Vert\mu\Vert\geq 0.5$ after early observations that benefit correlates with the mean-vector norm. \texttt{all-MiniLM-L6-v2} is the only deliberately-included low-$\Vert\mu\Vert$ model. The selection bias is disclosed in the body and quantified in Appendix~\ref{app:snow_share} for retrieval.

\section{Related Work}\label{app:related}

Pre-trained sentence embeddings exhibit \emph{anisotropy}: representations concentrate in a narrow cone whose dominant axis carries most of the variance. This was diagnosed early for word embeddings (a representation-degeneration problem during language-model training) by \citet{gao2019representation}, mapped explicitly for contextual embeddings by \citet{ethayarajh2019contextual}, and re-examined at the cluster-versus-isotropy level by \citet{cai2021isotropy}. Whether the resulting cone is the cause of poor semantic similarity, or merely a symptom of other systematic biases (token frequency, punctuation, etc.), has been debated by \citet{IsAnisotropy-2022} and revisited by \citet{IsotropyClusters-2024}, who argue that some anisotropy preserves cluster structure that downstream classifiers exploit. More recently, \citet{nastase2025geometry} showed that cosine similarity in sentence embedding spaces captures only shallow commonalities and does not reliably reflect how linguistic information is encoded, further motivating a careful, task-driven evaluation of any post-processing intervention rather than relying on cosine-based diagnostics alone. Our nine-method dose-response ablation is consistent with the latter view: removing \emph{all} anisotropy via full PCA whitening hurts every model we test, while a single-direction correction is small and reliably positive in the high-$\Vert\mu\Vert$ regime.

Several training-free or near-training-free post-processing techniques have been proposed to alleviate the cone. The closest ancestors of R1 and R2 are the smoothed-inverse-frequency baseline of \citet{SimpleBaseline-2017}, which removes the first principal component of an aggregated word-embedding average, and the All-but-the-Top algorithm of \citet{mu2017allbutthetop}, which centers the embedding distribution and removes its top principal components. ABTT-$1$ in particular is the dominant baseline for our R2 method: as we report in Section~\ref{sec:nine}, R2 and ABTT-$1$ are downstream-equivalent within $0.18$ pp on every model in our ablation despite weak geometric alignment between $\hat\mu$ and the centered top principal component (Appendix~\ref{app:pc}). We accordingly position R2 as a transparent, single-vector implementation of the same intervention rather than a new algorithm, with the additional contributions of an explicit error-propagation handle (Appendix~\ref{app:proof}) and the $\Vert\mu\Vert$ pre-deployment diagnostic (Section~\ref{sec:main}).

Whitening-style methods sit further along the dose-response axis. \citet{BERTFlow-2020} apply a normalizing flow to push BERT embeddings toward a Gaussian; \citet{WhiteningSentence-2021} and \citet{WhiteningBERT-2021} apply explicit decorrelation transforms. These are reported to improve sentence-similarity tasks in their original settings; on our ablation panel the most aggressive limit (full-rank PCA whitening on a corpus mean) hurts every tested model. We do not interpret this as a global refutation of all whitening-style approaches; rather, it documents a regime (modern multilingual sentence encoders with already-large $\Vert\mu\Vert$) in which untuned full-rank whitening is the wrong dose.

Beyond whitening, recent post-processing approaches include subspace manifold projection for anisotropy reduction~\citep{wren2025subspace}, attention-weighted re-projection (\texttt{Ditto}; \citealp{Ditto-2023}), large-scale dimension-drop ablations \citep{DimensionDrop-2025}, and LLM-driven embedding aggregation (\texttt{GenEOL}; \citealp{GenEOL-2025}). These act on different parts of the representation pipeline (token weighting, dimension subset, ensemble of paraphrases) than R2, which only edits the corpus-mean direction. Contrastive sentence-embedding training such as SimCSE \citep{gao2021simcse} also addresses anisotropy but at training time rather than as a plug-in correction. The Sentence-BERT framework of \citet{reimers2019sbert} gave the broader infrastructure on which most of the encoders we audit are built; the \texttt{e5} family of contrastively pre-trained encoders \citep{wang2022e5} is a representative recent backbone, and members of this family appear among our highest-$\Vert\mu\Vert$ audit subjects.

Finally, two recent papers use the shared mean direction \emph{as an attack surface} rather than as a quantity to remove: \citet{liang2025jailbreaking} construct universal magic-word attacks against text-embedding safeguards, and \citet{li2024zeroshotadversarial} use language-driven anchors as zero-shot adversarial directions on vision-language encoders. These works support the present paper's framing of $\mu$ as a coherent, model-intrinsic, transferable direction in the output space.

\section{Reproducibility Notes}\label{app:reproducibility}

\paragraph{Mean-vector estimation.} For every model, $\mu$ is estimated from the same $10^5$ sentences sampled from English Wikipedia (snapshot \texttt{20220301.en}), filtered to lengths in $[64, 512]$ characters. Sentences are pre-tokenized with NLTK; no task labels are used. The same corpus is used to fit centered PCA components for the All-but-the-Top and full-rank whitening baselines.

\paragraph{Whitening protocol.} For each model and task, full-rank whitening (a) centers all embeddings by subtracting $\mu$, (b) computes the SVD of the centered matrix, (c) scales each principal direction by $1/\sqrt{\lambda_i + \epsilon}$ with $\epsilon=10^{-6}$, and (d) $\ell_2$-renormalizes. PCA is fit once per model on the corpus described above and frozen across tasks.

\paragraph{Random-direction control.} The seed-$42$ random direction is sampled from $\mathcal N(0, I_d)$ then $\ell_2$-normalized. Multi-seed extensions on Snowflake-arctic-embed-m use seeds $\{2,3,4\}$ in addition. All seeds are documented in the supplementary release.

\paragraph{Effect-size and W/T/L thresholds.} The paired $t$-statistic is $\bar\Delta/\mathrm{SE}(\Delta)$ over per-task deltas within a (model, family) cell with at least $5$ tasks. Win/tie/loss counts use $|t|>2$ or $|\sigma_{\text{cell}}|>2$ as the threshold; this is an effect-size window, not a Bonferroni-controlled significance level. Bootstrap confidence intervals for Pearson correlations use $10\,000$ resamples with replacement.

\paragraph{Two MMTEB snapshots.} The $38$-model R1/R2 study was run on an earlier MMTEB snapshot (per-model task counts $609$ to $624$). The $5$-model nine-method ablation was run on MMTEB v$1.38.18$ (per-model intersections $333$ to $462$ across all nine methods). Per-row deltas in Table~\ref{tab:9method} are not numerically pooled with Table~\ref{tab:38model}.

\paragraph{Released artifacts.} Every paired per-task delta, every per-method per-task score, the per-model mean vectors, the multilingual-$\mu$ trajectory, the prompt-path probe outputs, the embedding-norm audit, the PC1-mean alignment files, the multi-seed random control output, and the per-study task-tracker metadata accompany this submission. Each main-body table, the headline figure, and every supplementary appendix table is regenerated by a corresponding script in the released supplementary materials.

\section{Compute}\label{app:compute}

The $38$-model R1/R2 audit was run on a single Linux host with several A6000-class GPUs. A typical model of size $\sim 100$\,MB requires roughly $24$ GPU-hours on one such GPU to complete the filtered MMTEB suite end-to-end. The $5$-model nine-method ablation was distributed across an H100-class cloud-GPU cluster. The cumulative GPU budget across the audit and the nine-method ablation is in the low single-digit thousand GPU-hours.

\end{document}